\documentclass[sigconf,nonacm]{acmart}
\AtBeginDocument{%
  }

\setcopyright{acmlicensed}
\copyrightyear{2026}
\acmYear{2026}
\acmDOI{XXXXXXX.XXXXXXX}
\acmConference[CHI '27]{Make sure to enter the correct
  conference title from your rights confirmation email}{May 10--14, 2027}{Pittsburgh, PA, USA}
\acmISBN{978-1-4503-XXXX-X/2018/06}

\usepackage[dvipsnames, table]{xcolor}
\usepackage{listings}
\lstdefinelanguage{Markdown}{
    basicstyle=\ttfamily\footnotesize,
}
\usepackage{enumitem}
\usepackage{graphicx}
\usepackage{subcaption}
\usepackage{caption}

\newcommand{\ilabel}[1]{
    \begin{picture}(7,8) 
      \put(3,3.5){\color{Dandelion}\circle*{11}} 
      \put(3.5,4){\makebox(0,0){{\textcolor{black}{\scriptsize\bfseries\sffamily #1}}}} 
    \end{picture}
}

\begin{document}

\title{Training Computer Use Agents to Assess the Usability of Graphical User Interfaces}


\author{Alice Gao}
\affiliation{%
  \institution{University of Washington}
  \city{Seattle, WA}
  \country{USA}}
\email{atgao@cs.washington.edu}

\author{Weixi Tong}
\affiliation{%
  \institution{Purdue University}
  \city{West Lafayette, IN}
  \country{USA}}
\email{tong172@purdue.edu}

\author{Rishab Vempati}
\affiliation{%
  \institution{Purdue University}
  \city{West Lafayette, IN}
  \country{USA}}
\email{rvempat@purdue.edu}

\author{Katharina Reinecke}
\affiliation{%
  \institution{University of Washington}
  \city{Seattle, WA}
  \country{USA}}
\email{reinecke@cs.washington.edu}

\author{R. Benjamin Shapiro}
\affiliation{%
  \institution{University of Washington}
  \city{Seattle, WA}
  \country{USA}}
\email{rbs@cs.washington.edu}

\author{Tianyi Zhang}
\affiliation{%
  \institution{Purdue University}
  \city{West Lafayette, IN}
  \country{USA}}
\email{tianyi@purdue.edu}

\author{Jason Wu}
\affiliation{%
  \institution{Purdue University}
  \city{West Lafayette, IN}
  \country{USA}}
\email{jasonwu@purdue.edu}

\renewcommand{\shortauthors}{Gao et al.}

\begin{abstract}
Recent HCI research leverages AI agents to assess graphical user interfaces~(GUIs) usability, but current agents still struggle. Computer use agents~(CUAs) that are used in these assessments are predominantly developed using objective task trajectories with verifiable outcomes and lack an operationalized definition of usability and necessary training data.
This work presents a novel training objective that operationalizes usability to fine-tune CUAs to assess GUI usability through relative comparisons between interface variants.
We apply our method to CUA models using a large-scale dataset of fully interactive UIs paired with usability labels mixed with human preferences. 
Two empirical evaluations across real and synthetic GUIs show that our method \textit{i)}~improved relative usability assessment accuracy across three off-the-shelf CUA models, \textit{ii)}~enabled our best model to outperform proprietary baselines, and \textit{iii)}~produced grounded usability assessments. 
Our work lays a data-driven foundation for agent-based usability assessment, enabling exploration of new forms of scalable usability evaluation and interface design.
\end{abstract}

\begin{teaserfigure}
    \centering
    \includegraphics[width=\linewidth]{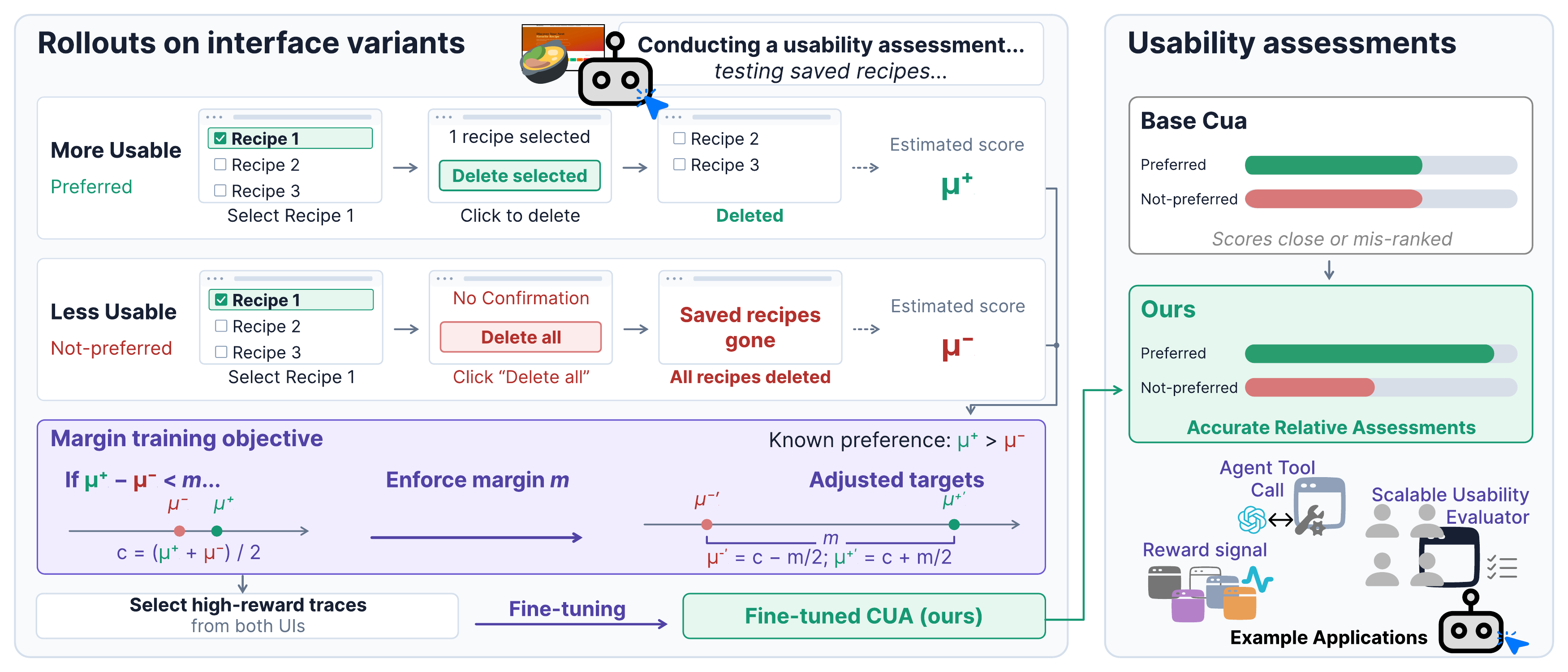}
    \caption{Our method fine-tunes computer use agents (CUAs) to perform accurate usability assessments by first collecting rollouts on interface variants and enforcing a margin training objective. The resulting CUA also has the potential to enable three HCI applications surrounding scalable usability evaluation and interface design.}
    \label{fig:teaser}
\end{teaserfigure}

\maketitle

\section{Introduction}
The concept of \textit{usability} captures how well an interface supports users in accomplishing their goals. Usability assessments, in turn, provide a basis for improving interface design. Human--computer interaction (HCI) has traditionally relied on varying and disparate empirical measures of usability through usability studies~\cite{nielsen1993numusers, dix2004human, findlater2007evaluating, mahatody2010state} and expert evaluations ~\cite{uxexpert, nielsen1990heuristic}. 
Without a standardized definition of usability, these traditional assessment strategies lack the ability of at-scale deployment.

Usability is difficult to operationalize as it is inherently subjective and largely dependent on individual experience. For example designers have shown large disagreements on which design issues users surface ~\cite{duan2024uicrit} and UI preferences vary depending on user groups. 
Even when following expert-written heuristics, individuals will vary in identified issues~\cite{nielsen1990heuristic}, as what may be a problem for one user may not even be a concern for another.
Prior HCI researchers have used computational models to predict how people will interact with interfaces, such as movement and task execution ~\cite{fitts1954information} or the perceptual, cognitive, and motor processes ~\cite{card1986model, anderson1998atomic} underlying interactions as a way to encode human notions of usability.
However, these models typically relied on explicit representations of the task, user, or environment, making them difficult to apply broadly across real-world interfaces.

AI agents powered by large language models (LLMs) offer a promising way to move beyond these limitations. Specifically, computer use agents (CUAs) are able to interact with a computer in a similar way as humans do \cite{guo2026susbench, gur2024realworldwebagentplanninglong}.
By combining broad pretraining with the ability to operate directly on rendered graphical user interfaces (GUIs), CUAs generalize across a much wider range of interfaces without requiring a custom model for each one. Recent HCI work has begun to use LLM-based agents for usability evaluations and related design tasks, though using diverse conceptions of usability. ~\citet{duan2024uicrit} used LLMs to offer design critiques through manually crafted guidelines, while others ~\cite{guo2026susbench, zhong2025syntheticheuristicevaluationcomparison, holter2026uxcascadescalableusabilitytesting, lu2025uxagent} have used them to conduct usability tests. However, unlike earlier models grounded in empirical observations of human interaction, CUAs are trained on large-scale data that are curated or synthesized without explicitly modeling real human behavior or usability. 
This can lead to hallucinations or incorrect predictions of human behavior ~\cite{lu2026llmagentssimulatemultiturn, zhong2025syntheticheuristicevaluationcomparison, gui2023}.

To align CUAs for GUI usability assessments, 
we propose to operationalize usability through \textit{relative comparisons} using synthetic interface variants. One way to scale usability metrics and provide interaction data needed to fine-tune CUAs is by manipulating an interface to inject defects, as the variant is now \textit{less} usable. 

We contribute to data-driven assessments of UI usability by developing a novel training objective for CUAs to more accurately assess GUIs. In that vein, we fine-tune CUA models using relative comparisons of interface variants. To support this training, we create uxWeb, the first large-scale dataset of 2,586 fully interactive GUIs paired with usability labels (plain vs defect-augmented) mixed with human usability judgments. Labels are obtained through a synthetic augmentation pipeline that instructed coding agents to inject common usability errors into websites by modifying their source code. We use our method and dataset to train three off-the-shelf CUA models to improve their usability assessment capabilities (e.g., overall models should assess UIs with known defects as \textit{less usable}).

We conduct two separate evaluations to validate our approach. First, we evaluate whether our method improves usability assessment across three off-the-shelf CUA models, finding consistent improvements (22.6\% in AUC) over their corresponding base models. We also compare our best-performing model against strong proprietary and open-source baselines, finding that it outperforms much larger models by 18.1\%. Second, we show that improvements in usability assessments resulting from our method were more grounded in observations from their navigation traces on real and synthetic GUIs. We discuss how our method can be used in downstream applications such as scalable simulated usability testing, agent-assisted user interface design, and reward signal for generative UI techniques.

To summarize, we make the following contributions:
\begin{enumerate}[noitemsep,topsep=0pt]
    \item We create and will publish uxWeb, a large-scale dataset of fully interactive clones of original websites and their defect-augmented counterparts, annotated with usability labels and judgments by designers.
    \item We introduce a novel method for fine-tuning computer use agent (CUA) models to make accurate usability assessments and show that it provides consistent improvements over corresponding base models, enables our best-performing model to outperform larger proprietary models, and is accompanied with more grounded usability assessments on real and synthetic websites.
\end{enumerate}

\section{Related Work}
Our research builds on existing work surrounding UI usability evaluation and user simulation by encoding human notions of usability into computational models. We contextualize our work by providing an overview of UI usability evaluation, user simulation, and computer-use agents.

\subsection{User Interface Evaluation}
Traditionally, UI evaluation has relied on experts to conduct usability studies with target users~\cite{usability101} or heuristic evaluations using a set of guidelines~\cite{nielsen1990heuristic}.
While expert-driven evaluations can offer rich and targeted feedback, these approaches are costly and only capture the opinions of a small number of evaluators, remaining difficult to scale.
As a result, crowdsourced approaches or ``arenas'' have been used to incorporate feedback from larger numbers of non-experts~\cite{luther2015crowdcrit} and are widely used to benchmark the design performance of AI models on design tasks~\cite{designarena_website_2026}. These approaches enable more rapid feedback, enabling designers to improve UIs for the user experience. Other solutions have focused on modeling human behavior~\cite{oulasvirta2018aalto, gajos2005preference, gajos2004supple} to assess and adapt UIs.

Simultaneously, UI evaluation is not limited to generalized usability; other types of UI evaluations exist aimed towards improving specific aspects of usability. One particular focus is to determine and taxonomize dark patterns~\cite{digeronimo2020uidark, gunawan2021darkpatterns}, or UIs intentionally design to mislead users. Others threads of research have focused on detecting design defects~\cite{yang2021don, wu2024uiclipdatadrivenmodelassessing} or accessibility issues~\cite{fok2022accessibility, zhong2025screenaudit} and even eliciting design preferences that may impact usability for different user groups. Our work extends existing work in user interface evaluation, aiming towards providing automated usability assessment in a data-driven manner.

\subsection{User Simulation}
One approach of improving automated UI evaluation is to accurately simulate users of those interfaces.
Early work in HCI and cognitive psychology approximated UI usability by combining cognitive and physical measurements (e.g., reaction time and motor speed) with interaction traces to approximate time needed to complete tasks.
ACT-R~\cite{anderson1998atomic}, incorporated richer models of human cognition, such as memory, attention, and learning to more realistically simulate these user-interface interactions.
Together with interaction and task models~\cite{paterno1997concurtasktrees}, these user models have been employed to automatically design and evaluate UIs, through approaches collectively known as model-based UI development~\cite{puerta1997model}. However, these approaches largely relied on hand-crafted models and manually specified rules rather than data-driven learning of user behavior.

Since LLMs and other foundation models are trained on large volumes of human-written text and artifacts, researchers hypothesized these models to be effective simulations of human responses and interactions~\cite{park2022socialsimulacracreatingpopulated,hamalainen2023evaluating,wang2026interview}, also applying these approaches to simulate UI feedback.
Recent work has explored prompting LLMs with UI screenshots, personas, and design guidelines to simulate comments from end-users~\cite{xiang2024simuser} and experts~\cite{duan2024uicrit,zhong2025syntheticheuristicevaluationcomparison}.
Since usability evaluation often depends on interface interaction rather than static screenshots, researchers have also developed LLM-based agents and crawlers capable of navigating and operating applications as another method of simulating users.
Computer use agents (CUAs) have been used to test the functionality of GUIs and identify bugs through ``crawling''~\cite{yu2025visionbased,gao2026guitester}.
Another avenue has been to incorporate these systems into evaluation protocols, such as automated judges~\cite{li2026webdevjudgeevaluatingmllmscritiques} or supporting expert-driven evaluation~\cite{holter2026uxcascadescalableusabilitytesting,lu2025uxagent,zhong2025screenaudit}.

LLM-based user simulation relies on the assumption that LLMs are accurate models of users, but several works have shown mismatches in human ratings~\cite{shankar2024validates}, responses~\cite{kapania2025simulacrum}, and assessments~\cite{wu2024uiclipdatadrivenmodelassessing}.
This suggests a need to incorporate additional training objectives or additional data to refine off-the-shelf models. 
\citet{oulasvirta2022computational} theorized that human users behave as bounded rational agents, and their behavior can be approximated by reinforcement learning training objectives.
Subsequent work applied these techniques to learn user simulations for typing~\cite{shi2024crtypist}, VR selection~\cite{ikkala2022breathing}, and menu navigation~\cite{moon2023amortized}.
Others have fine-tuned LLMs using human data such as feedback~\cite{duan2024uicrit,wu2024uiclipdatadrivenmodelassessing,zheng2023judging} and interaction traces~\cite{deng2023mind2web} so that their behavior more closely aligns with real users.
Our work combines both approaches to fine-tune CUAs for usability testing, with the goal of better aligning their assessment scores and navigation behavior with human users.

\subsection{Training Computer Use Agents}
Foundation models are promising for UI understanding because of their pretraining on large amounts of web data and text gives them strong capabilities for interpreting interfaces and user instructions.
Many approaches for building CUAs prompt foundation models with representations of a target UI and along with the model's context window and a goal.
CUAs can rely on text-only LLMs by converting UIs into intermediate representations (e.g., pseudo DOM) to execute actions~\cite{wang2023enabling, deng2023mind2web} or they can use multimodal models~\cite{openai2023gpt4v}, where they integrate visual prompting techniques to ``ground'' their actions.
While prompting methods can be effective, they rely heavily on the pre-existing capabilities of the underlying foundation models.

To address this issue, later work has explored learning-based approaches that use demonstration trajectories to train or fine-tune CUAs.
For example, Mind2Web combined manually annotated web trajectories with fine-tuned action prediction models~\cite{deng2023mind2web}, while WebGUM more directly framed web navigation as offline multimodal training over a large corpus of demonstrations~\cite{furuta2023multimodal}.
Other systems such as UI-TARS and OpenCUA trained end-to-end GUI agents on large action-trace datasets and human computer-use demonstrations~\cite{qin2025uitarspioneeringautomatedgui,wang2025opencua}, and provided automated judges to assess the quality of web development~\cite{li2026webdevjudgeevaluatingmllmscritiques}.
Another learning-based approach to improving CUA performance is by designing new training objectives or applying reinforcement learning in online environments.
Examples include AndroidEnv, which provides an interactive mobile environment for agent training~\cite{toyama2021androidenv}, and more recent methods such as DigiRL, WebRL, and EvoCUA, which improve agents through online interaction, reinforcement learning, or large-scale synthetic rollouts~\cite{bai2024digirl,qi2025webrl,xue2026evocua}.
As a result, most CUAs are trained for task completion, e.g., following a user's natural language command. 

\section{Dataset of Interface Usability}
To explore a data-driven approach for usability assessment capabilities of machine learning models, we first describe our approach to building uxWeb, a large-scale dataset of 2,586 fully interactive websites paired with usability labels and human judgments.
Since a large-scale dataset with the location and severity of real-world usability flaws is not readily available, uxWeb is constructed through a combination of synthetic replication of popular websites, principled data augmentation, and human expert ratings.

\subsection{Synthetic Websites}
uxWeb consists of sets of synthetically generated website clones with usability labels created from synthetic augmentation and human labeling. We describe the \textit{i)} website generation process to create \textit{plain clones} of existing sites, \textit{ii)} synthetic usability augmentations to create a set of \textit{defect-augmented} counterparts for each plain clone, and \textit{iii)} our protocol for collecting human ratings.

\subsubsection{Website Generation}
We first generated clones of popular websites drawn from Mind2Web, a standardized web agent navigation benchmark~\cite{deng2023mind2web} that contains 137 websites, across 5 high-level domains (e.g., shopping, entertainment) and 30 subdomains.
Generating standalone ``clones'' of websites made them easier to deploy, allowed synthetic control of website behavior (e.g., through code edits) and reduced barriers to agent use (e.g., login credentials). 

To create these clones, we used the Claude Code agent~\cite{anthropic_claude_code_docs_2026} configured with GLM 4.7~\cite{zai_glm47_modelcard_hf_2025} as the backend endpoint for cost efficiency. At the time of our experiments, Claude Code was the best performing coding agent, and GLM 4.7 was the best-performing open source code generation model~\cite{designarena_website_2026}.
We provided GLM 4.7 a bare bones React.js template, to avoid requiring the model to generate ``boilerplate'' configuration code.
The coding agent was then prompted to replicate an existing site from Mind2Web based on its name and a description of its functionality. It was also prompted to ensure that the website clone was fully functional, self-contained, and with mocked backend flows that represented the real website as closely as possible.
Placeholder assets such as images were retrieved using an online API that was queried with model-generated alt-text annd mock data that was self-generated by the model. The full prompt is provided in Appendix \ref{app:website-generation}. Because many of the websites are popular, the coding agent was often able to reconstruct their general interaction flows and common tasks from prior knowledge alone, without requiring screenshots of the reference site. Our goal was not to produce clones that matched the original websites exactly in terms of functionality or appearance, but instead to generate a diverse distribution of realistic websites for our experiments.
Once website generation finished, the model also created a txt file documenting all the user flows it had implemented. We used this file for quality control, checking if the first few clones generated had realistic and representative user flows implemented before continuing to generate the rest of our dataset. We also used this file when injecting usability defects. Each website clone took 5-7 minutes for GLM 4.7 to fully implement. The 879 websites required approximately 102.5 machine-hours to generate, which was distributed over the course of a month.

Since synthetic data generation can introduce hallucination, such as the model claiming to have implemented a user flow it did not, or introduce defects due to probabilistic outputs, we cloned each of the 137 Mind2Web sites eight times. We performed two additional rounds of quality checks resulting in a total of 879 plain website clones. The first involved creating a script that would check if the website could be rendered in Playwright, a headless Chrome browser. If the resulting screenshots were were blank, the website was removed from our dataset. The second quality check occurred during our training procedure, which we discuss in \S\ref{sec:trace-heuristics}. 

\subsubsection{Usability Defect Injection}\label{sec:usability-defects}
Like their real-world counterparts, our synthetically-cloned websites were not free of usability defects but the type, location, and severity of these defects remains unknown for both real-world sites and our synthetic ones.
To support modeling and benchmarking, we developed an approach to inject \textit{known} usability defects into each of our plain clones in a programmatic manner to create sets of \textit{defect-augmented} versions for each plain synthetic clone. This augmentation process allowed us to measure model performance on a known (but possibly incomplete) list of usability flaws. This process also resulted in a \textit{relative} ranking of website usability between \textit{plain} and \textit{defect-augmented} sites, which we later used to calibrate absolute score prediction through our training objective.

To design these usability defects, we reviewed various usability and design guidelines~\cite{appleguidelines, usability101, 5principlesvisualUX, lidwell2010universal}. 
Ultimately, we chose to adapt Ben Shneiderman's 8 Golden Rules of Interface Design~\cite{shneiderman2010designing} for their relevancy to our task and ability to operationalize\footnote{These principles also have significant overlap with Nielsen's Heuristics~\cite{usability101}.}. Below, are the eight types of usability defects inspired by these principles: 
\begin{itemize}[noitemsep,topsep=0pt]
    \item \textbf{Consistency}: Consistent sequence of actions should be required in similar situations (e.g., identical terminology used in prompts, menus, help screens, consistent layout, capitalization, fonts).
    \item \textbf{Feedback}: For every user interaction there should be interface feedback. Feedback should be proportional to frequency and importance of actions.
    \item \textbf{Dialog}: Dialogs should yield closure and should have a clear beginning, middle, and end.
    \item \textbf{Prevention}: The interface should be designed so that a user cannot make serious errors; erroneous actions should leave the interface state unchanged.
    \item \textbf{Control}: Users should feel as if they are in control of the interface; they should not get unexpected surprises or have tedious data-entry sequences.
    \item \textbf{Reversal}: As much as possible, actions should be reversible.
    \item \textbf{Memory}: Avoid designs that require users to remember information when moving from one display to another. 
    \item \textbf{Hierarchy}: Important elements central to the task should be emphasized and easy to find.
\end{itemize}

We automated our usability defect injection in a similar manner to our initial website generation. For each existing plain clone, we prompted our coding agent to inject code changes to introduce defects that violated one of our usability principles above. We provided GLM 4.7 the codebase for the existing website and the usability principle to violate, then instructed it to introduce changes that would violate the user flows documented in the generation step. The coding agent was instructed to keep all changes contained in the \texttt{src} folder and not introduce changes that would break website functionality. 
We injected usability defects into 388 plain sites. Usability defect injection took 7-10 minutes per principle, taking approximately 284.5 hours to generate our 1,707 defect-augmented sites. The prompt for usability defect injection is in Appendix \ref{app:website-generation}. 

To validate our usability defect injection resulted in \textit{i)} sites that were less usable than the plain counterpart and \textit{ii)} contained defects corresponding to the injection type (i.e. usability principle), we adopted the procedure from ~\citet{zhang2025slideaudit} on a subset of uxWeb. Three authors blindly inspected 5 plain sites and a random sample of their defect-augmented counterparts for a total of 47 sites. We found that for 75\% of sites, there was unanimous agreement that defect-augmented sites were less usable. For 73\% of the sites, there was at least two-thirds agreement that the type of flaws the defect-augmented site contained was the same as the category that was injected.

uxWeb was randomly partitioned into training, validation, and testing splits based on unique website name. We reserved one website per each unique subdomain (30 total) for our test set. 15\% of the remaining websites were used for validation, resulting in 92 unique websites and 15 unique websites for the training and validation splits respectively. In total we had 1,573 clones for training, 199 clones for validation, and 393 clones for testing.

\subsection{Designer-Rated Usability}\label{sec:designer-pref}
Our synthetic approach supports the scalable generation of potentially large numbers of training examples, which is beneficial for model training. However, the usability defects injected to create our defect-augmented sites may not capture usability flaws that were present in the original website clones or reflect human perception of usability, limiting the generalization of models trained on this data.
Therefore, we collected an additional split of data labeled by human designers.

\subsubsection{Designer Rating Procedure}
To collect designer ratings of these clones, we constructed a \textit{model arena}. Model arenas are interfaces commonly used for LLM feedback collection and benchmarking~\cite{designarena_website_2026,lmarena2025webdevarena}, and they require participants to rate multiple candidate outputs for the same input prompt. In our case, we developed a feedback interface where two plain websites \textit{without} intentionally-introduced defects corresponding to the same source website and prompt were compiled and displayed side-by-side in \texttt{iframe} elements, allowing experts to use the websites while preserving interactivity. The interface also contained a button that required users to expand the \texttt{iframe} for more detailed inspection and access the website using the full resolution of their browser window.

We recruited 17 participants at university level or higher with design background (i.e. university courses or work experience) to partake in our study to rate 30 preference pairs. After navigating the functionality of each site, participants were asked to provide a response that indicated which of the two sites were more usable. Participants were instructed that the sites were clones of existing sites and that placeholder data was used. Participants were recruited through word of mouth or university Slack channels. Our study protocol was determined as exempt by our institution's IRB. Study details can be found in Appendix \ref{app:study}. 

In total, we collected 510 preference pairs, which spanned 428 unique plain website clones. Using server logs, we found that participants spent an average of 54.83\footnote{We removed outliers based on z-score with $z>3$ in cases where the participant may have left the study tab open before coming back to complete a preference rating.} seconds on each pair and clicked on the \texttt{iframe} window an average of 4.65 times and on an average of 3.05 elements before their final response. This suggests that participants spent effort in testing website flows before making assessments.
Nevertheless, when we computed inter-rater agreement for a subset of pairs with multiple labels, we found an agreement rate of 0.308 (Krippendorff's $\alpha$). We expected moderate-to-high levels of disagreement, as the websites being compared were all generated from the same model and same prompt, and were therefore more likely to contain similarities. This also mirrors results found by other prior research~\cite{duan2024uicrit,wu2024uiclipdatadrivenmodelassessing,wu2025improving_ui_generation_designer_feedback,peng2025designprefcapturingpersonalpreferences,luther2015crowdcrit}, which found high levels of subjectivity in design-related evaluation. 
\section{Training Approach}
In this section we provide an overview of our novel training objective to train computer use agents to \textit{i)} prioritize exploration of important UI flows and \textit{ii)} and accurately predict usability scores that match ground-truth judgments. We also describe how we use this approach to train CUAs and our agent formulation.

\subsection{Problem Formulation}
Our resulting CUAs are designed to perform GUI usability assessment in the same way humans do, through visual perception and clicking. While research has shown that having access to hidden semantic information like the HTML or DOM allows web agents~\cite{browseruse, gur2024realworldwebagentplanninglong, furuta2024multimodalwebnavigationinstructionfinetuned} to more effectively find information and executing tasks, we chose to formulate the CUA to operate more generally on any type of GUI, using the same visual reasoning a human would typically undergoe (e.g., item search). Based on its initial observation of the input UI, the CUA predicts interaction flows that are important to overall usability (e.g., room booking for a hotel site) and prioritizes important user flows to check within a given interaction budget (e.g., 50 actions). After the navigation phase, the CUA reviews its interaction session and reasons about the UI's usability strengths and weaknesses, then produces a numerical value (0-100) representing a usability score. This was to ensure the model provided a bounded assessment.
We choose this method as over rubric-based scoring due to typical coverage deficiency and noisy evaluations from LLM-based scoring \cite{thakur2025judging, tan2024judgebench, shen2026rethinkingrubricgenerationimproving}, and the fact that no comprehensive usability scale exists.
This score was initialized by prompting the model to estimate the probability a user could successfully complete the agent's tested flows given the observed design defects. The score was calibrated and further refined using a training objective that enforces a margin (i.e., minimum difference) between preferred and rejected UIs in our dataset.

\begin{figure}
    \centering
    \includegraphics[width=\linewidth]{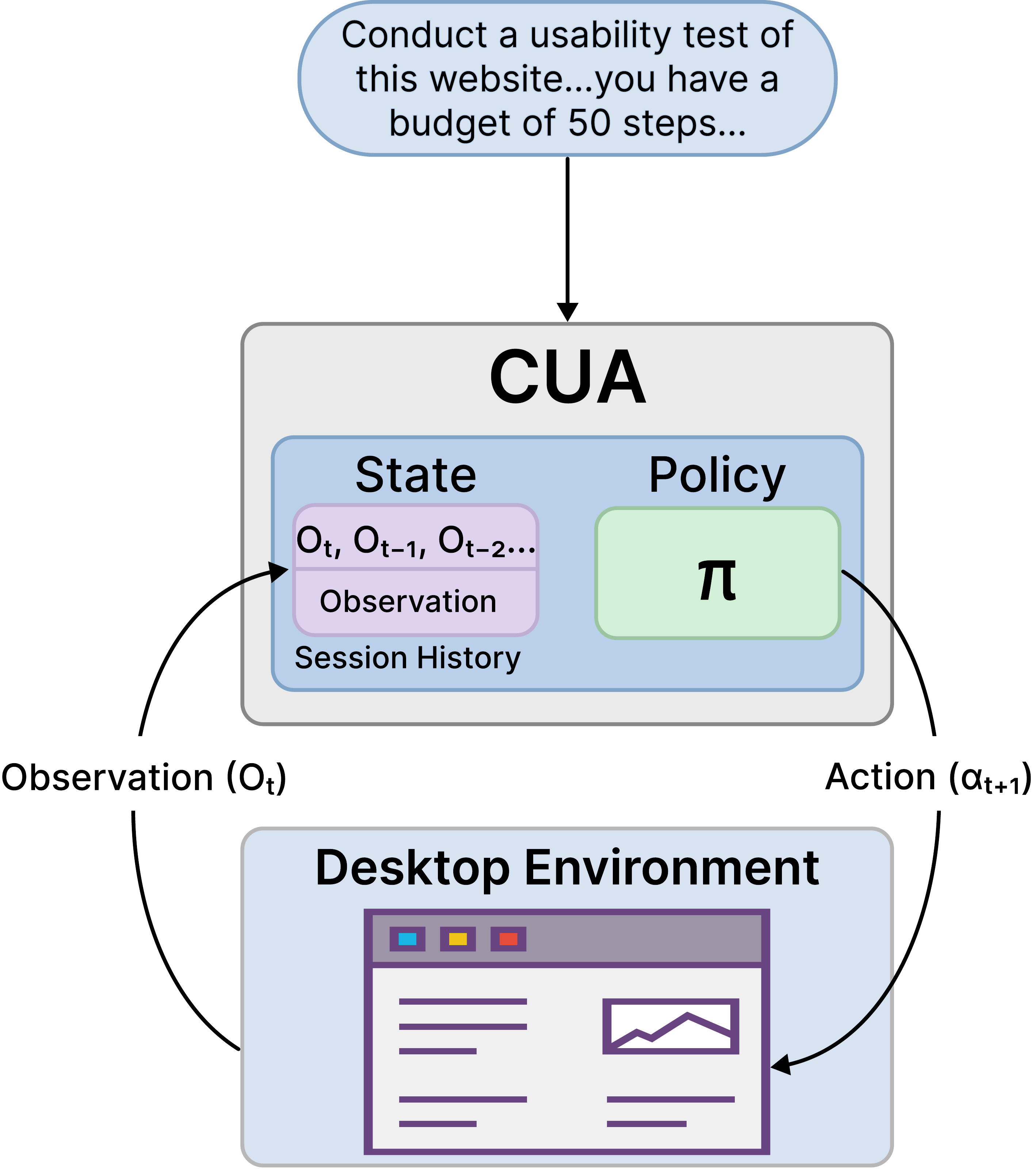}
    \caption{Our CUA architecture for GUI usability assessment. At each timestep, the agent receives an observation $o_t$ in the form of a screenshot of GUI running in a headless desktop environment, which it uses to update its internal state $s_t$ before executing an action $a_t$ under its current policy $\pi$.}
    \label{fig:model-arch}
\end{figure}

\subsection{Agent Architecture}
Our CUAs are based on a popular open-source CUA agent harness~\cite{qin2025uitarspioneeringautomatedgui} that captures a stream of screenshots from the user's desktop, feeds them to the underlying model hosted on a vLLM inference endpoint~\cite{kwon2023efficient}, and uses generated responses to directly control the keyboard and mouse.
This allows the CUA to run on any graphical desktop environment (e.g., operating system) and interact with any GUI regardless of the underlying implementation.

The CUA architecture follows the default settings of UI-TARS and maintains several data structures.
First, a textual goal, $G$, provided to the agent at the beginning of execution.
While most CUAs are trained to perform well-defined tasks (e.g., navigating to a web address), we use a static usability testing prompt (Appendix \ref{app:crawl}) that instructs the model to make its own determination of important interaction flows to test for the purposes of rating usability.
At each time step, the harness captures a visual screenshot, or observation $o_t$, which to update the agent's internal state $s_t$, which consists of a window of the five most recent visual observations and a JSON-like representation of the session history.
$s_t$, $o_t$, and a grounding instruction is used to prompt the the CUA model to reason about how to make progress towards $G$ and describe observations (similar to a ``think-aloud'' exercise) to produce an executable action $a_t$. Valid actions are defined by the UI-TARS harness and exposed to the model through an initial system prompt. A single step of execution and processing lasts approximately five seconds.
The agent stops its navigation when a special ``stop'' action is generated, e.g., if the model believes it has finished exploration, or if a maximum number of time steps (50) was reached.
The final trace reflection and usability scoring is implemented by appending a manually-crafted evaluation prompt (Appendix \ref{app:crawl}) as an additional turn in the session history and generating a response using the CUA. We provide an overview of this architecture in Figure \ref{fig:model-arch}.

\subsection{Training Algorithm}\label{sec:training-algo}
\begin{figure*}[!hptb]
    \centering
    \includegraphics[width=\linewidth]{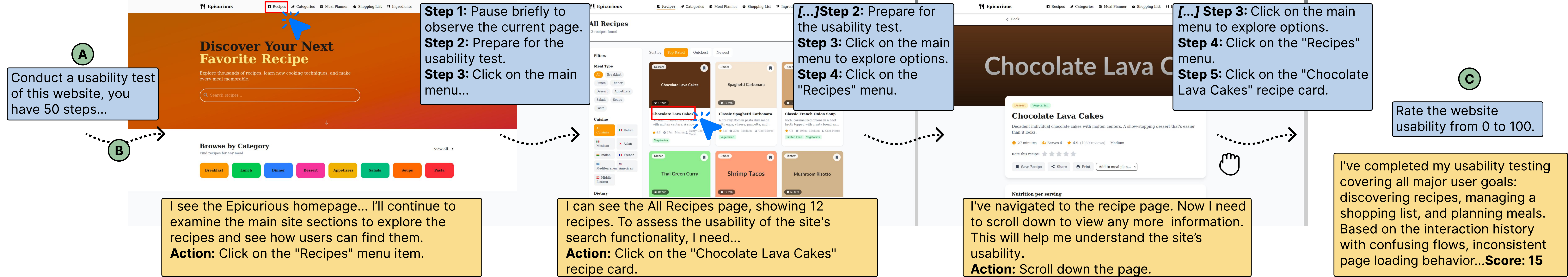}
    \caption{A portion of an interaction trace that the model executes on a site from uxWeb. At the beginning (a) the model is asked to conduct a usability test; (b) the model starts exploration through an identified flow; (c) the model concludes its exploration and is prompted to assess the site's usability based on its interaction history.}
    \label{fig:trace-window}
\end{figure*}

We adopted an approach based on existing policy-based RL training pipelines \cite{ahmadian2024basicsrevisitingreinforcestyle, williams1992simple} to train the CUA to effectively \textit{i)} explore UI flows, \textit{ii)} test their functionality, and \textit{iii)} assess their usability through a score.
For our experiments, we chose the 8B or 9B versions of the best-performing open-source models on the OSWorld benchmark~\cite{xie2024osworld_neurips} as they were the strongest models that could fit on our hardware at the time of experiments.

In this section, we describe the three stages of our training approach: \textit{i)} rollout generation, \textit{ii)} reward assignment, and \textit{iii)} policy update.

\subsubsection{Rollout Generation}\label{sec:trace-generation}
In policy-based reinforcement learning approaches, training data is obtained by running the current policy in the environment and collecting rollouts that reflect its present behavior, later adjusted through learning.
In our experiments, we configured our agent harness with the base model (i.e., EvoCUA-8B, GUI-Owl-1.5-8B, Qwen3.5-9B) and generated up to 24 trace rollouts for each website in the training split of uxWeb by repeating its automated usability test protocol.

To perform rollout generation, we developed a containerized Apptainer environment that contained \textit{i)} a headless X display server running at 1080p resolution, \textit{ii)} an instance of Firefox running in full-screen mode, and \textit{iii)} our agent harness which controlled a virtual mouse and keyboard.
We wrote a script that repeatedly sampled a website from the training split at random, which was a React.js application bundle. This bundle was transferred to the container, where the website was staged and hosted at a pre-defined local URL. After successful staging, it was loaded into the browser, and a new usability testing session was started using our CUA harness.
The CUA harness was configured to query a vLLM instance running on a compute cluster that served the base model.
We configured our harness to log all interactions, model responses, and screenshots to accurately represent the states and actions of the rollout. Figure \ref{fig:trace-window} shows a visualization of a portion of a collected rollout. 
We generated an average of 26,989 traces per model, taking approximately 2.5 weeks (machine hours) per model. Prompts for trace generation and scoring are provided in Appendix \ref{app:crawl}.

\subsubsection{Reward Assignment}\label{sec:trace-heuristics}
After rollout generation, we evaluated and assigned rewards based on the \textit{i)} quality of the navigation trace and \textit{ii)} accuracy of the agent's predicted usability score. Rollouts that are assigned high reward scores are used to amplify effective behaviors within the model's policy.

\paragraph*{Navigation Quality}
The quality of navigation behavior has a large impact in usability assessment, and CUAs can often exhibit behaviors that are not characteristic of human users, such as failure to click graphical targets or ``looping'' between a small number of screens ~\cite{xue2026evocua}.
We developed several heuristic measures to validate the navigation quality of rollouts used in training.

For each rollout, we computed the following metrics:
\begin{itemize}
    \item \textbf{Same screen ratio:} The fraction of transitions with no visual change. Higher values indicate more repetitive behavior.
    \item \textbf{Same screen after clicks ratio:} The fraction of clicks that produced no screen change. Higher values suggest ineffective clicks or that the CUA became stuck.
    \item \textbf{Unique screen ratio:} The fraction of unique visited screens. Higher values indicate broader website exploration.
    \item \textbf{Steps:} The total number of steps in the interaction trace. Some traces were shorter due to early failure.
\end{itemize}
We filtered out blank screens (possibly due to load failures) before computing the first three metrics as they could artificially reduce these scores. We used perceptual hashing (phash) \cite{phash} with a Hamming distance of 4 to compute whether two screens were the same. Perceptual hashing has applications in reverse image search and identifying similar screens \cite{agten2015seven, hao2021phash, kintis2017hiding, rajasegaran2019multi}. 
We then combined these three metrics to compute the navigation quality score, $s_{nav}$: 
\begin{equation}
\small
    \begin{split}
    s_{nav} = \text{Unique screen ratio} &\times (1 - \text{Same screen ratio}) \\
                   & (1 - \text{Same screen after clicks ratio})
\end{split}
\end{equation}\label{eq:quality_score}
We computed $s_{nav}$ for unique screens in the rollout that were not separated by scrolling actions, which could trigger false positives in the screen hashing metric. We thresholded this score and rollout length to filter our generated rollouts. Through trial and error we determined that a $s_{nav}$ of at least 0.07 and a minimum of 30 steps indicated a quality trace. 
We had an average of 3,973 traces per model that passed this quality check and used to train our models.

As an additional filter for \textit{website} quality, we excluded websites in uxWeb if fewer than 3 of its 24 rollouts passed our heuristics, since we hypothesized that they were likely to contain implementation errors that prevented useful rollouts to be collected.

\paragraph*{Usability Assessment Accuracy}
After navigation, the CUA must accurately reflect on its observations, reason about the UI's strengths and weaknesses, and provide a numerical usability score.
Assigning a reward that reflects the accuracy of a rollout's usability assessment is difficult due to the lack of ground truth for both the design reasoning and a usability score.

A straightforward but likely incorrect approach is to use a 100 as the target usability score for positive samples (plain websites) and 0 as the target usability score for negative samples (defect-augmented websites).
Yet, the absolute quality of websites in our data is not known, e.g., usability defects may exist outside of the ones that were intentionally introduced, which would make 100 the incorrect target score.

\begin{figure}[!hptb]
    \centering
    \includegraphics[width=\linewidth]{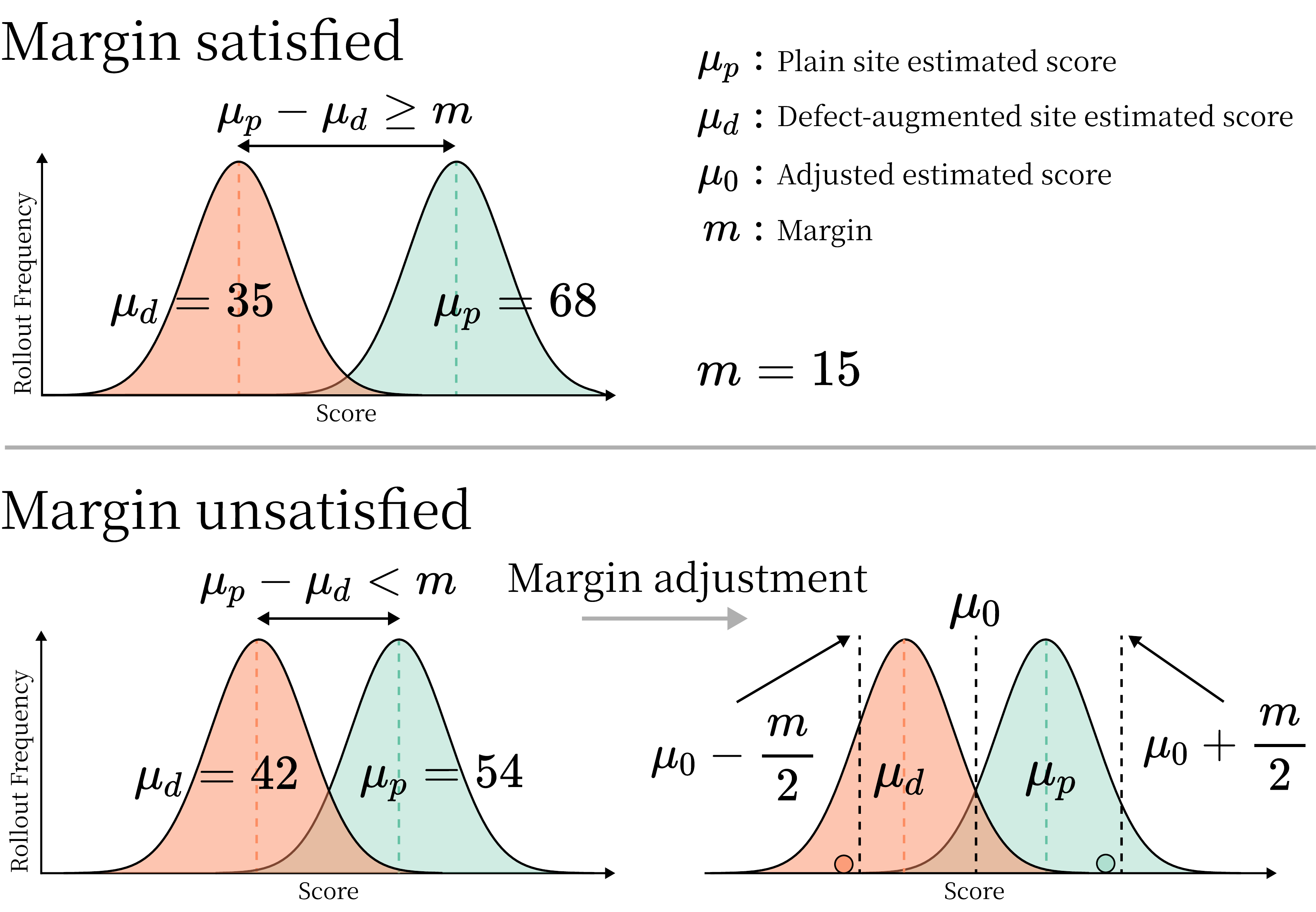}
    \caption{If the estimated target scores for plain sites ($\mu_p$) and defect-augmented sites ($\mu_d$) satisfy the margin we use their estimated scores. Otherwise, we enforce a margin by shifting their estimated scores.}
    \label{fig:target_score}
\end{figure}

To address this challenge, we introduce a method for dynamically estimating the target usability score of a website based on a model’s initial score predictions for both the website and its defect-augmented counterpart.
Our approach, inspired by other contrastive learning objectives for text responses~\cite{,lambert2026rlhfbook,touvron2023llama2} and visual design quality~\cite{wu2025improving_ui_generation_designer_feedback}, enforces a margin between model-predicted scores of preferred and non-preferred samples (Figure \ref{fig:target_score}).
We first estimate the score of the plain website, $\mu_p$, as the mean predicted usability score across all of its rollouts, and use the same procedure to estimate the score of the corresponding defect-augmented website, $\mu_d$. If the estimated score difference does not satisfy the desired margin, i.e., $\mu_p - \mu_d < m$, we define adjusted target scores such that the positive website is assigned a target of $\mu_0 + \frac{m}{2}$ and the negative website is assigned a target of $\mu_0 - \frac{m}{2}$, which are the values closest to the initial model estimates that satisfy the margin and with $\mu_0 = \frac{\mu_p+\mu_d}{2}$ .
Reward values are then assigned to rollouts for each website based on the proximity of their predicted usability scores to the corresponding target score, such that rollouts with predictions closer to the target receive higher reward.
We chose the value of the margin $m$ based on manual tuning, and we include it as a hyperparameter in the Appendix.

\subsubsection{Policy Update}
Finally, we used our dataset of reward-labeled rollouts to update the CUA's policy model.
For each site in the training split of uxWeb, we selected a single rollout (out of 24) with the highest assigned reward, which led to a total of 631 rollouts.
We converted each rollout into multiple training examples using a sliding-window approach, where each training example contained the expected input and output format of the policy network at each timestep in the rollout~\cite{qin2025uitarspioneeringautomatedgui,xue2026evocua}.
We generated an average of 21,228 windows per model for fine-tuning our policy.
Examples from all websites in the uxWeb training split were aggregated for training.
We used LLaMAFactory~\cite{zheng-etal-2024-llamafactory}, an open source library, to conduct supervised fine-tuning using 
reward-labeled rollouts.
We employed 4-bit quantization and LoRA to improve training efficiency. Hyperparameters were determined through manual experimentation and provided in Appendix \ref{app:hyperparameters}. 
Model fine-tuning was conducted on a GPU compute node with 4 H100 GPUs.
\section{Technical Evaluations}
We conducted two evaluations to assess models trained with our objective.
Our first evaluation compared usability assessment accuracy, or the relative usability scoring, for models trained with our objective to that of existing open-source and proprietary CUA models. 
Our second evaluation assesses models' intermediate usability reasoning during their navigation traces on synthetic and real websites through the quality of their usability critiques to see if more accurate assessments are accompanied with grounded observations.

\subsection{Evaluation 1: Usability Assessments}\label{sec:usability-scoring}
We compared the accuracy of usability assessments of CUAs trained with our objective, CUAs trained with a naive objective, their open-source base models, and proprietary models. We evaluated the accuracy of model-predicted assessments using labels from uxWeb, which included both synthetic and human-provided labels.

\subsubsection{Baselines}
We used the following multimodal models as baselines for usability assessment:\begin{itemize}
    \item \textbf{GPT-5-mini} \cite{gpt5mini}: OpenAI's proprietary efficient reasoning model for coding, computer use, and agentic workflows. GPT remains one of the leading choices for proprietary models and at the time of our experiments, this was one of the latest models released.
    \item \textbf{Kimi K2.5} \cite{kimiteam2026kimik25visualagentic}: Moonshot AI's open-source 1.1T-parameter reasoning model for coding and computer use. At the time of our experiments, it ranked first among open-source models and second overall on OSWorld~\cite{xie2024osworld_neurips}.
    \item \textbf{EvoCUA} \cite{xue2026evocua}: Meituan's open-source CUA model. Its 8B variant (EvoCUA-8B-20260105) was the strongest OSWorld open-source model that could run on consumer hardware at the time of our experiments.
    \item \textbf{GUI-Owl-1.5} \cite{xu2026mobileagentv35multiplatformfundamentalgui}: Tongyi's lab open-source CUA model. This model ranked second after EvoCUA on OSWorld. We selected its 8B variant. 
    \item \textbf{Qwen3.5}: Alibaba's most recent open-source model. At the time of our experiments, Qwen3.7 Plus ranked \#3 on OSWorld but was not open-sourced. We selected this model as an alternative and its 9B variant.  
\end{itemize}

\subsubsection{Model Training and Objectives}
We followed the training approach from \S\ref{sec:training-algo} for each open-source model (EvoCUA, GUI-Owl-1.5, Qwen3.5). We trained each open-source model with our margin objective and performed an ablation with a naive 0/100 target objective, replacing the target usability scores of plain sites with 100 and defect-augmented sites with 0. 

\subsubsection{Methodology}
We first constructed test samples, i.e., preference pairs, from the test split of uxWeb.
From our test split, we randomly selected one plain clone per unique source website (30) and randomly sampled 4 of 8 of the plain clone's associated defect-augmented variants for a total of 150 sites.
We also collected all designer ratings of websites in the test split. In total, the benchmark consisted of 150 unique uxWeb website implementations and 115 comparison pairs.

During our evaluation, all CUAs were configured with the same harness and settings, which standardized their input/output modalities, i.e., visual observations of the GUI and keyboard/mouse interactions (\S\ref{sec:trace-generation}), and generated chain-of-thought reasoning logs during their interactions.
All conditions were run in our container-based GUI execution environment that staged websites locally and launched them in a full-screen Firefox browser. One interaction trace was generated for each evaluated site, limited to a maximum of 50 timesteps. 
Next, we applied each model to each site in a preference pair with the model's prediction computed by taking the higher usability score of the sites in preference pairs. In cases where the model did not provide a score, we counted that as a vote for the other site. 

We used \textbf{AUC} (area under precision-recall curve) as our primary metric to assess the degree to which the model's predicted score discriminated between chosen sites (e.g., plain sites or human-preferred) over rejected ones (e.g., defect-augmented or human-rejected), compared to random chance. AUC accounts for the class imbalance in the four-to-one ratio of defect-augmented to plain websites in the test set.

\subsubsection{Results}
\begin{figure*}[!hptb]
    \centering
    \includegraphics[width=0.8\textwidth]{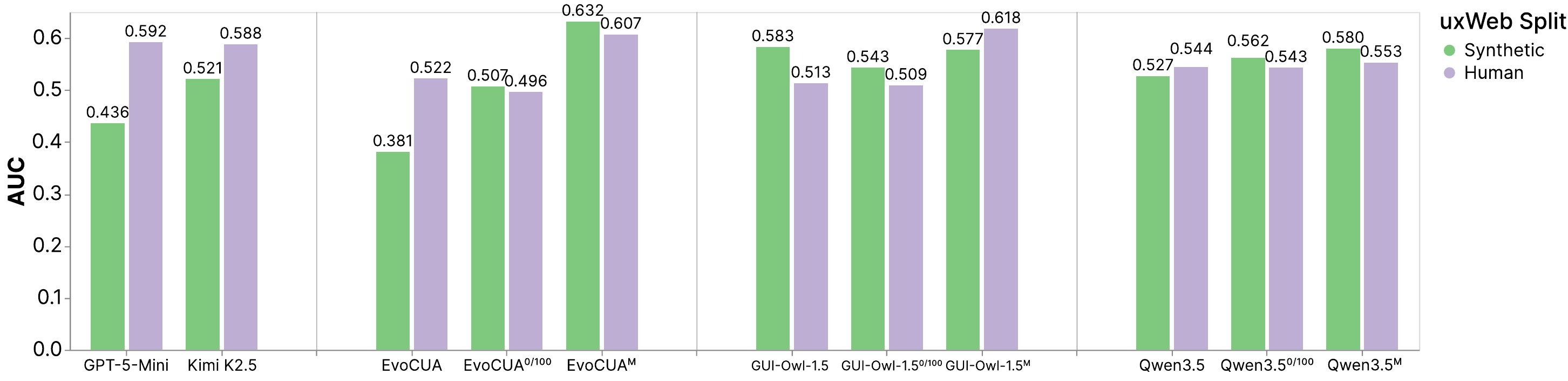}
    \caption{Training with our margin objective leads to overall improvements over base models in AUC compared to a naive 0/100 objective. Our best performing model (EvoCUA$^{M}$) even outperforms proprietary models in providing accurate usability assessments for both test splits.}
    \label{fig:auc}
\end{figure*}

Overall, our training objective resulted in models with increased accuracy in usability assessments (Figure \ref{fig:auc}). Our margin objective lead to a consistent performance increase over base models (22.6\%) compared to a naive 0/100 objective, which resulted in minimal improvement (8.0\%). Additionally, our best fine-tuned model outperformed proprietary models by 18.1\% on both splits.

Without training, \textit{baseline models provide inaccurate usability assessments on UIs with known usability issues.} We observe how baseline models provide inaccurate usability assessments for defect-augmented sites (Figure \ref{fig:auc}), with assessments worse than chance (GPT-5-mini, EvoCUA) or only slightly better than chance (Kimi K2.5, Qwen3.5, GUI-Owl-1.5). In contrast, our training objective results in models that can outperform their base models by 32.4\% and proprietary models by 25.6\%. Our best fine-tuned model demonstrates notable improvement in predicting usability assessments to sites with known usability issues, with an AUC of 0.632. This model provides a 37.0\% increase in AUC over base and proprietary models, as well as a 65.9\% increase over its base model (EvoCUA).

Incorporating training on human preference pairs with our objective also lead to overall improvement (12.8\% increase) over base models. Similar to before, AUC for baseline models remains low (Figure \ref{fig:auc}). Their agreement rate with what humans label as more usable UIs suggests results comparable to random guessing. As described in \S\ref{sec:designer-pref}, the original agreement rate between designers was low ($\alpha$=0.308) so large improvements may be difficult. Our best fine-tuned model achieves an AUC of 0.607.

Overall, our results suggest that our training paradigm with our data is an effective method for learning relative usability scoring, providing resulting models the ability to both form accurate usability assessments on sites with known defects and usability assessments aligned with human judgments.  

\subsection{Evaluation 2: Usability Critiques}\label{sec:reasoning-analysis}
Our second evaluation investigates whether the improved assessment capability resulting from our training objective was grounded in accurate observations from its model's navigation trace. To do this, we examined the intermediate steps and reasoning generated by the model while the CUA performed its tasks and the final reflection phase used by the model before generating its final usability score. Since this requires manually inspection of traces, we focus this evaluation to the best model (EvoCUA$^M$) from \S\ref{sec:usability-scoring} and its base model (EvoCUA).

\subsubsection{Methodology}
To evaluate the model's internal critiques, we adopted the procedure from \citet{duan2024uicrit}, which was used to evaluate the performance of a UI screenshot critic pipeline. We chose not to apply \citet{nielsen1990heuristic}'s heuristic evaluation protocol since it primarily depends on overlap and recall of issues identified by human-experts, who do not ``hallucinate'' like models, i.e., requires measuring precision. Following existing protocol, a critique could be \textit{i)} valid, \textit{ii)} partially valid, or \textit{iii)} invalid. We adapted the definitions to be applicable to interactions: a critique is considered \textbf{valid} if it exists on the site and was identified by the CUA during its navigation, \textbf{partially valid} if it does not exist on the site but resulted from the CUA's trace (i.e. the CUA made a correct observation based on its trace but would not be an issue for humans such as misclicks due to time delay in harness which may have also resulted in the incomplete execution of a user flow), and \textbf{invalid} if it does not exist and the CUA did not encounter the issue during its navigation trace (hallucination). Critiques were assigned 1.0, 0.5, and 0 points respectively. Three authors manually examined traces from 10 randomly selected plain benchmark sites of each model along with traces on corresponding real sites for a total of 20 sites. Authors were blinded to the model during annotation to reduce the chance that their ratings would be influenced by knowledge of the model that generated it. In total 251 issues were examined. 

\subsubsection{Results}
\begin{figure}[!htbp]
\centering
\begin{minipage}[t]{\linewidth}
\centering
\footnotesize
\captionof{table}{Our training objective results in a model (EvoCUA$^M$) is also accompanied with overall better usability critique quality, or identified usability issues based on the model's navigation trace, compared to its base model (EvoCUA).}
\begin{tabular}{llc}
    \toprule
    \textbf{Split} & \textbf{Model} & \textbf{Avg. Usability Critique Quality Score} \\
    \midrule
    Synthetic & EvoCUA & 0.44 \\
     & EvoCUA$^M$ & 0.62\\
    \midrule
    Real & EvoCUA & 0.37  \\
         & EvoCUA$^M$ & 0.39\\
    \bottomrule
\end{tabular}
\label{tab:usability-issues}
\par\vspace{3em}
\end{minipage}
\begin{minipage}[b]{\linewidth}
    \centering
    \includegraphics[width=\textwidth]{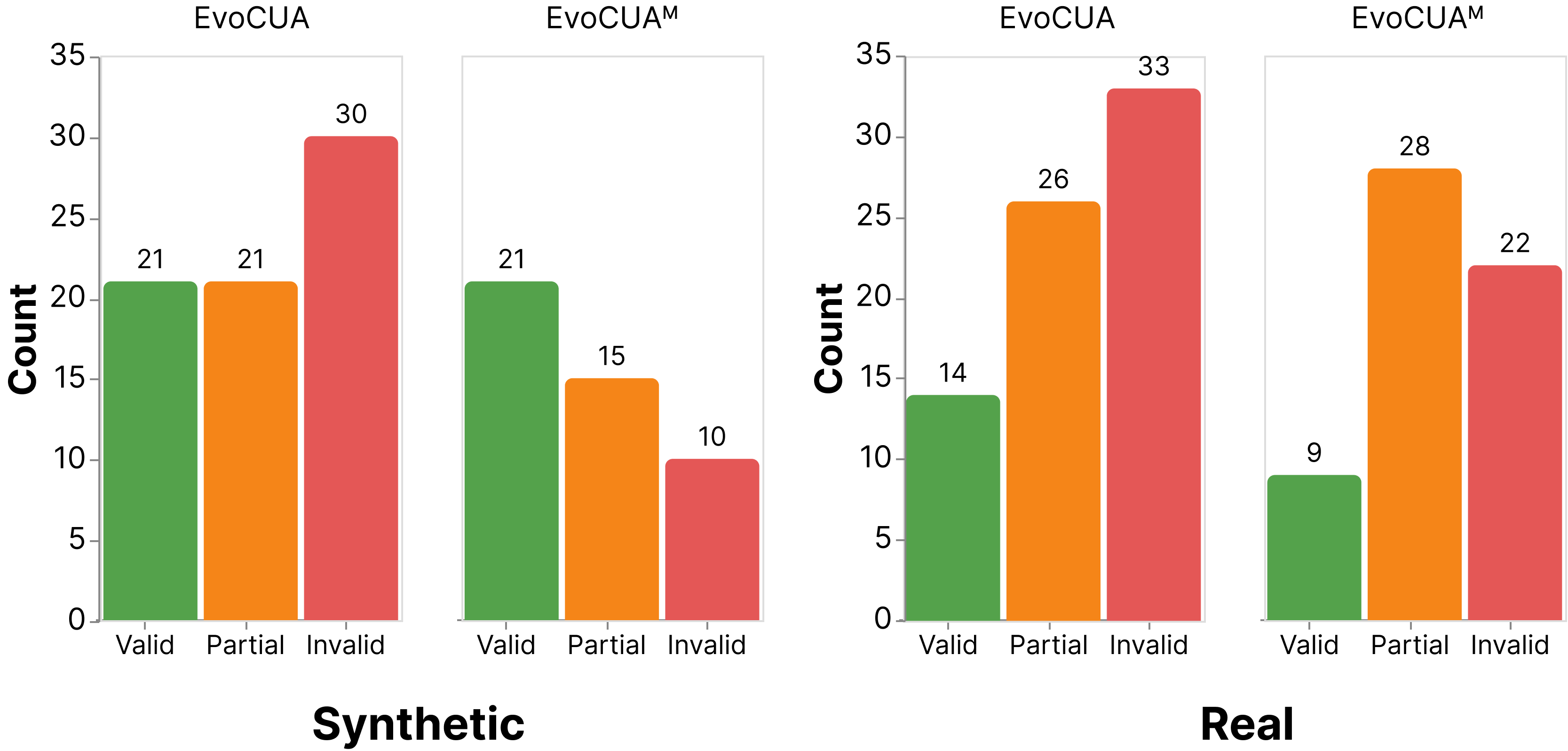}
    \captionof{figure}{Fine-tuned on our objective, EvoCUA$^M$ overall produces less partially valid and invalid critiques on both synthetic and real websites, suggesting its increased usability assessment accuracy was grounded in its navigation traces.}
    \label{fig:usability-issue-count}
\end{minipage}
\end{figure}

Overall, we observed an average 23.1\% increase in usability critique quality on real and synthetic sites. Table \ref{tab:usability-issues} shows the average usability critique quality, as calculated by the procedure introduced by~\citet{duan2024uicrit} . We achieved similar inter-rater reliability score (Krippendorff's $\alpha=0.35$) as ~\citet{duan2024uicrit} (Fleiss' $\kappa=0.29$) on critique quality ratings.

Our fine-tuned model produced fewer partially valid critiques and fewer invalid critiques (Figure \ref{fig:usability-issue-count}), suggesting a decrease in the number of hallucinated observations. Despite gains in both domains, the fine-tuned model had a smaller increase on real sites and lower overall performance.
This suggests that our method is primarily effective in improving model performance on in-domain websites and usability defects but also leads to limited improvements on out-of-distribution data.
On real-world data, the tuned model had a larger number of partially valid critiques.
Upon manual inspection, we noticed that on real sites, models sometimes attempted to perform tasks that required login credentials then encountered verification issues that were interpreted by the model as usability faults. We marked these as partially valid, since they were faithful to the model's real observations but would not constitute a problem for a human user. Overall, results from this experiment suggest that the better usability assessments previously observed are accompanied by more accurate grounding.

\section{Example Walkthrough}\label{sec:walkthrough}
\begin{figure*}[!hptb]
    \centering
    \includegraphics[width=\textwidth]{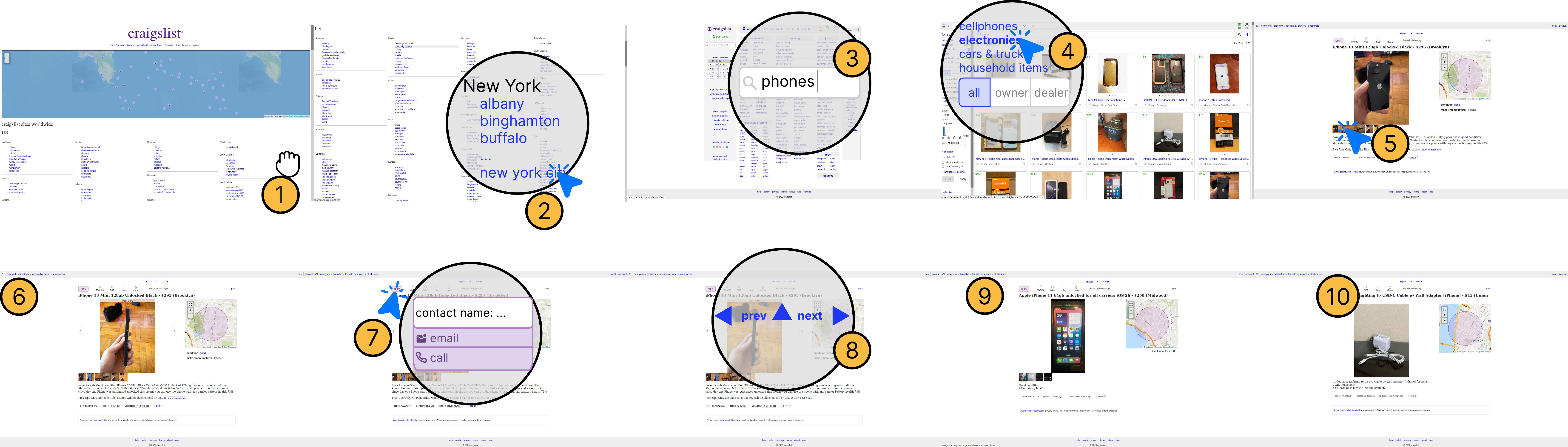}
    \caption{An abridged interaction trace as the model conducts a usability test on craigslist.org. In this example, the agent chooses to explore flows related to searching for items on sale.}
    \label{fig:uxcua_cw}
\end{figure*}

In this section, we provide an informal walkthrough\footnote{For the purpose of this walkthrough we provide abridged versions of turns.} as our model navigates a real-world site, demonstrating that it provides a usability assessment grounded in its interaction history.

We selected \textit{craigslist.org} and provide an example walkthrough as our model navigates the website. We highlight areas where our model successfully and unsuccessfully explores the website and provide its final usability assessment as an example (Figure \ref{fig:uxcua_cw}). 

Our CUA begins when we instruct it to conduct a usability test on craigslist.org. The agent first waits for the page to load before beginning \ilabel{1}): 
\begin{lstlisting}[language=Markdown,escapeinside={(*}{*)}]
    (*\textbf{Agent}*): I'll start by identifying the most important user goals for this site. Craigslist is primarily used for posting and searching for classified ads, so I'll need to test these flows.
\end{lstlisting}

Observing the list of US states and cities, the agent decides to explore local listings in New York City (\ilabel{2}), reasoning that is a major city that will enable it to better test the site's features. After navigating to New York City Craigslist page, the agent decides to test the functionality of the search box with a common item it believes to be most useful to users, a phone (\ilabel{3}).

\begin{lstlisting}[language=Markdown,escapeinside={(*}{*)}]
    (*\textbf{Agent}*): I see the page has loaded with various iPhone-related items. The search results are for "iphone" but the category filter on the left is set to "cell phones." I'll adjust the filter to "electronics" to refine results and test functionality.
\end{lstlisting}

Once the agent successfully refines its search (\ilabel{4}), it decides to test flows for browsing and examining product details. To do so, the model then clicks on a result for an iPhone, spending a few turns clicking through the image gallery(\ilabel{5}, \ilabel{6}). Once it finishes testing the gallery function, it progresses to see if it can view additional details about the phone or seller's contact information. After verifying that the seller's contact information can be viewed (\ilabel{7}), it continues its usability test and attempts to navigate back to the main electronics listings page.

Here(\ilabel{8}) the agent decides to click on the \texttt{prev} button at top of the screen to return to the main listing page, but ends up on a different iPhone listing instead. This is likely due to the fact that the way to navigate back to Craigslist's search results page is actually through the uparrow ($\uparrow$) between the \texttt{prev} and \texttt{next} hyperlinks and not the hyperlinks themselves. At this point the agent, unable to navigate back to the listings page, decides to continue its current interaction flow by exploring more iPhone listings and examining each iPhone listing in detail to test the functionality of details views before it reaches the maximum number of steps (\ilabel{9}, \ilabel{10}). At the end, the agent provides its final usability assessment: 

\begin{lstlisting}[language=Markdown,escapeinside={(*}{*)}]
    (*\textbf{Agent}*): I see issues with: 
    - (*\textbf{Navigation}*): Repeatedly click on "prev" and "next" buttons to navigate content suggests inconsistent or unreliable navigation, indicating potential issues.
    - (*\textbf{Confusing Flows}*): Interaction history shows the assistant needs to scroll multiple times to view content, suggesting the layout or content organization may be unintuitive. 
    - (*\textbf{Failed Page Loads}*): There were no page load failures, which is good. 
    Based on the these issues, the website usability score is (*\textbf{65}*).
\end{lstlisting}

Overall, we observe that the model is able to successfully navigate and identify key flows it should test on Craigslist, knowing how to search for items, identifying inconsistencies, and to explore page details. 
At the end of its exploration, the agent provides as grounded assessment based on its interaction history.

\section{Discussion}\label{sec:discussion}
We introduced a novel training objective to train CUAs to assess GUI usability and show that through our approach, we can outperform baselines and even beat proprietary models in accurate usability assessments, which is accompanied with more grounded usability assessments for real and synthetic UIs. We next discuss key results and broader implications.

\subsection{Computational Definition of Usability}
A central goal of this work is to define usability assessment in a form that can be learned computationally. Usability depends on multiple factors, including aesthetics, functionality, and how easily users can discover important functionality. The UI perception and navigation capabilities of the underlying foundation model are important, but insufficient on their own. For example, our experiments found that models with stronger OSWorld navigation performance could navigate sites but lacked grounded usability assessments, frequently assigning websites with known defects usability scores akin to chance.

One way to operationalize usability is to prompt models with design guidelines~\cite{zhong2025syntheticheuristicevaluationcomparison} or expert-written critiques~\cite{duan2024uicrit}. However, these approaches often produce invalid or only partially valid critiques, likely because positive examples alone do not discourage hallucinations and a model's retrieved examples may come from mismatched contexts.
Another approach is to supervise the evaluation process itself by training on expert traces. In practice, this is difficult: different evaluation methods can produce different findings~\cite{gray1998damaged}. Additionally traces are often noisy, and there is no single canonical path to a correct usability judgment. Such data is also hard to augment synthetically.

We therefore focused on outcome supervision rather than direct process supervision~\cite{lightman2023letsverifystepbystep,shao2024deepseekmath}. Outcome labels are easier to generate at scale and less costly for humans to provide. Prior work has done this for static screenshots~\cite{wu2024uiclipdatadrivenmodelassessing}, but interactive usability assessment makes process and outcome tightly coupled: incorrect judgments may result from either poor exploration or poor reasoning. Our work is an initial step toward encoding an HCI evaluation procedure into a model, and we hope similar approaches can help formalize and learn other HCI methods.

\subsection{From Better Models to New HCI Applications}
Our approach builds on prior HCI work on user simulation for usability testing with the goal of enabling new, scalable HCI applications. To evaluate their final products, designers typically conduct usability studies where they provide participants with tasks~\cite{usability101}, so simulating users means modeling how they respond to predefined stimuli or goals~\cite{card1986model,anderson1998atomic,zhang2025artifactsbench}. They then synthesize those results, weighing the severity and tradeoffs especially when comparing alternatives. As the web development cycle becomes increasingly rapid and agentic \cite{lu2025uxagent, holter2026uxcascadescalableusabilitytesting}, feedback must match the pace of development. We envision our method and resulting models as a way to provide designers early-stage evaluations before incorporating human-centered feedback as well as integrated into the automated design tools they use. We outline these applications below. 

\subsubsection{Scalable Simulated Usability Testing}
Scalable simulated usability testing becomes increasingly necessary as the number of interfaces needing evaluation outpaces traditional usability testing. Our method can be used to train models that can provide these on-the-fly evaluations. Trained models can assess the usability of candidate interfaces and their alternatives, such as during A/B testing, generated by designers. These relative predictions can be repeatedly collected then aggregated before presented as a final score to designer or as a histogram of scores. As these assessments are made, usability critiques across runs can similarly be aggregated then grouped by issue type. Designers can then view the final score(s) and critiques for candidate interfaces to make judgments about the severity of the issues or to prioritize which issues to fix. They can fix flagged issues and improve the interface before the process is repeated, or the site is deployed/tested by real users. This enables rapid iteration before involving human feedback and may reduce usability issues in earlier stages of the development cycle.

\subsubsection{Embedded Tool Call for Agents}
Vibecoding platforms have seen a rise in popularity as they enable both designers and general users to create fully-interactive websites with a single prompt. Though current general purpose LLM services (e.g. Claude Artifacts \cite{claude_artifacts}, ChatGPT Canvas \cite{chatgpt_canvas}) or specialized vibecoding platforms for website creation (e.g. Lovable \cite{lovable}, Replit \cite{replit}) have seen significant progress in website generation, the resulting sites are not necessarily usable. Often, they lack core functionality and beyond the first few webpages \cite{uxdesign_cc_exposing, mop_genui_fail} as they are primarily used in initial stages for planning. Similarly, coding agents (e.g. Codex, Claude Code) are also limited in assessing usability of generated UIs. Generated UIs may be visually pleasing but lack functionality \cite{wu2024uicoderfinetuninglargelanguage} as agents fail to understand how visual adjustments impacts the UI's usability \cite{jeon-etal-2026-mllms}. We envision our method and resulting trained models integrated into these existing agents as an \textit{agent tool}. After the agent (on the platform or coding agent) generates a UI, they can then perform an \textit{agent tool call} to our models for usability assessment. Multiple usability assessments can be made for the agent to iteratively improve the interface before returning a final interface to the end-user.

\subsubsection{Reward Signal for Improving UI Generation}
We also envision that our method can be used to directly improve models that generate UIs and are integrated into current platform services or coding agents. Currently, generated UIs are limited to landing pages and small-scale sites with limited functionality \cite{uxdesign_cc_exposing, mop_genui_fail}. RLHF can be used to steer underlying models towards generating full-fledged, usable sites. However, existing RLHF methods used to improve generators ignore the rationale used to critique UI designs and improve their usability \cite{wu2026improving}. To enhance existing LLMs, a fine-tuned model with our method can first be used to assess generated UIs. The resulting relative usability assessment scores can be used by existing ranking-based RLHF methods to rank these UIs, enabling models to learn which UIs are more usable.  

\subsection{Limitations \& Future Work}\label{sec:limitations}
While we show our method can outperforms other baselines in our experimental data, we see several avenues for improvement.

\subsubsection{Technical Limitations}
Our CUAs are constrained by both compute limits and current CUA capabilities. Crawls were capped at 50 steps, likely preventing complete coverage of a website. This was due to practical constraints and by design: our training objective emphasized prioritizing important tasks rather than exhaustive exploration. Accordingly, unlike a traditional crawler, our CUAs were not intended to achieve full coverage in this initial work. Additionally, our CUAs are slow, requiring at least five second per step when accounting for the agent's internal timing and model inference on our hardware. Due to this delay, they sometimes makes errors that humans typically would not (in \S\ref{sec:walkthrough}), such as missing click targets and misclassifying them as usability issues. However, our best performing CUA's large improvements over its original model (EvoCUA) and performance compared to proprietary models still suggest that our training objective can be applied to more powerful models and improve their usability assessments. With less computational limits, future work can also explore applying our training approach with video streams to improve model usability assessment performance.

The model was trained primarily on fixed-resolution websites, due to the available synthetic data and to the training environment of the base model~\cite{xue2026evocua}. We expect this approach to transfer to similar 2D WIMP interfaces, but likely less well to mobile settings. Finally, our website evaluations did not include strong guardrails against high-impact actions, such as 
making purchases or deleting data. Prior work highlights the importance of accounting for the consequences of risky or irreversible UI actions~\cite{zhang2025interactionimpact}, which was one reason we limited evaluation on real-world sites.

\subsubsection{Generalizability}
Our CUAs were trained and primarily evaluated on synthetic websites, limiting its generalizability. We hypothesize that with access to real world interface variants, such as alternative UIs produced during A/B testing, or datasets collected by large scale LLM benchmarks (e.g. LM Arena), model performance would improve in these domains as our evaluations show overall in-domain improvements. We used synthetic sites because they allow controlled manipulation of usability characteristics, reproducible evaluation, and terms-of-service-compliant data collection without requiring user accounts or payment information. However, they are less polished than real websites and may not capture the full range of real-world usability issues. Prior work has shown that CUAs can be influenced by dark patterns, suggesting that some interfaces may be harder for both our model and humans to evaluate~\cite{guo2026susbench}. Although our walkthrough provides some evidence that our CUAs can operate on real websites, our findings may not extend to all intended use cases, such as other other GUI types (e.g., native desktop or mobile apps) or intermediate stages of a designer's workflow. Still, the strong gains over EvoCUA-8B on both splits of the synthetic dataset suggest that our architecture and training approach are effective when suitable data is available, and may transfer to other domains with sufficient domain-specific data.

\subsubsection{Contextualized Evaluation}
Finally, we note limitations of our evaluation and directions for future work. We focused on outcome decision and critique accuracy, but these do not fully capture the process by which our CUAs reaches its assessments. Even when the model matches human preferences, it is unclear whether this is because it explores the same parts of the interface or relies on similarly meaningful evidence. 

Different use cases will also require different capabilities and evaluation criteria. Our approach focused on training CUAs to provide accurate usability scores as a relative assessment for a site's usability, in other scenarios such as designer judging, direct designer agreement may be more appropriate. Similarly, in other contexts, designers may value actionable feedback and control over the tasks, instructions, and context provided to the model. Practitioners using our CUAs in model training may need interpretable and calibrated scores. In end-user or generative UI settings, the model may also need to adapt to individual preferences and accessibility needs. We hope our released dataset and models provide a foundation for these more targeted applications and evaluations.
\section{Conclusion}
We present a novel training objective to train CUAs to assess GUI usability. We used it to fine-tune different open-source base models. To support this goal, we created uxWeb, a large-scale dataset of interactive websites annotated with both synthetic defect-based usability labels and human judgments. Across two evaluations, our method results in consistent gains over base models, enables our best-performing fine-tuned model to outperform larger proprietary models, and is accompanied with grounded usability critiques on real and synthetic interfaces. These findings suggest that current off-the-shelf agents are not (yet) sufficient for rigorous usability assessment, but that usability assessment can be learned when paired with appropriate data and objectives. More broadly, we hope this work provides a foundation for scalable, reproducible, and human-aligned usability evaluation in HCI, and that releasing our data and models will support future research on automated design evaluation, benchmarking, and interactive AI systems.

\bibliographystyle{ACM-Reference-Format}
\bibliography{refs}


\appendix

\section{uxWeb Creation Prompts}\label{app:website-generation}
\subsection{Synthetic Website Generation}\label{app:website-generation-prompt}
We used the following prompt for GLM 4.7 when generating websites clones of the 137 real-world sites from Mind2Web. This prompt was determined by trial and error and manual inspection on a small number of test cases.
\begin{lstlisting}[language=Markdown]
    You are an expert web developer and designer specializing in modern websites. You are currently handed a barebones React.js template and your jobs is to make a fully working website using React.js. Feel free to use lightweight libraries like Tailwind CSS to enhance the design.
    
    Requirements:
    1. Create a fully functional, modern, and responsive website design
    2. Use only React.js, HTML, CSS, and JavaScript, but feel free to use libraries like Tailwind CSS to make the design better and make layout easier. You can look at the package.json file to see which libraries are already installed and import them. However, you must NEVER install new packages that were not already installed.
    3. Include interactive elements where appropriate
    4. Make it production-ready and professional
    5. You must include all relevant script tags for libraries to work properly.
    6. There should NEVER be any dead links or buttons that are visible but do nothing when clicked.
    7. The ONLY time you are allowed to run the `npm` command is to test if the website builds with `npm run build` command. You must NEVER run `npm` under any other circumstances.
    8. Because you will create a website with very realistic mocked functionality, you will probably need some data. ALWAYS synthetically generate placeholder data that is needed by any functionality on the mocked website. NEVER attempt to connect to online APIs.
    9. NEVER include images for which you do not have the appropriate path to (e.g., the "src" attribute). You should use an API to generate a placeholder instead like <img src="https://placehold.co/600x400" alt="Placeholder image">
    10. Do not define any new SVGs - you should use icons from FontAwesome instead from the packages: @fortawesome/fontawesome-svg-core @fortawesome/free-solid-svg-icons @fortawesome/react-fontawesome. You can assume that these packages are installed even though you cannot see them in a package.json in the same directory.
    11. NEVER attempt to access any external directory outside of your current directory.
    
    Your job is to create a fully functioning clone of the following website: <WEBSITE_NAME>
    
    Use the current barebones React.js project with React Router to implement this website, with all important functionality realistically mocked. The main React file is located at src/App.jsx and exports a component called "App." Make sure your code conforms to this structure. Your workflow should be something like the following:
    
    1. Think about the target website. Think about as many flows as you can that a user would want to do on this website.
    2. Implement the React.js website so that all of flows you thought are fully implemented, interactive, and using the appropriate mock data.
    3. Run `npm run build` command to test that website compiles without any errors.
    4. Return to step 1 and repeat until you cannot think of any more flows.
    
    When you think you are done, scan for any dead links or unimplemented functionality and implement them.
    
    Finally, when you are done, make a txt file in the root directory of the React.js project (at the same level as the src/ directory) called "flows.txt" that contains all of the flows that you implemented.
\end{lstlisting}

\subsection{Usability Defect Injection Prompt}\label{app:usability-defect-prompt}
We used the following prompt to inject usability defects for our defect-augmented websites. For each website clone, we used this prompt up to eight times to inject usability defects aligned to one of our principles with the provided definition in \S\ref{sec:usability-defects}. 
\begin{lstlisting}[language=Markdown]
    You are an expert web developer and UI/UX designer in charge of usability testing. You will be given a completely 1) a completely functional website and 2) potential user flows that can be performed on the website. The website is implemented with React.js with React Router. Your job is to introduce code changes that violate a specific interaction design principle, making the app less usable for that princeiple, while not breaking the website. 
    
    Types of changes you can make: 
    1. Delete: // deleting UI elements 
    2. Add: // including additional UI elements 
    3. Modify: // modifying UI elements and their functionality
    
    App: <WEBSITE_NAME>
    Flows: flows.txt
    Principle: <PRINCIPLE>
    
    
    Using the current React.js code in the src/ directory: 
    1. Consider all user flows inside the .txt file and all, the critical steps to these flows, and what changes can be introduced to violate the given principle 
    2. Change the code in the current directory to violate the given principle. Ensure these changes are fully implemented, interactive, and use the appropriate mock data or modified mock data
    3. Run `npm run build` command to test that website compiles without any errors.
    4. Run `npm run build` to check if all changes made in step 2 are implemented AND show up in the modified code
    
    When you are done, make a text file in the current directory (src/) called changes.txt listing all the changes made to the application that have made it more difficult to use.    
\end{lstlisting}
\subsection{Defect-Augmented Examples}
\begin{figure*}[!ht]
    \centering
    \begin{subfigure}[b]{0.6\linewidth}
        \centering
        \includegraphics[width=\linewidth]{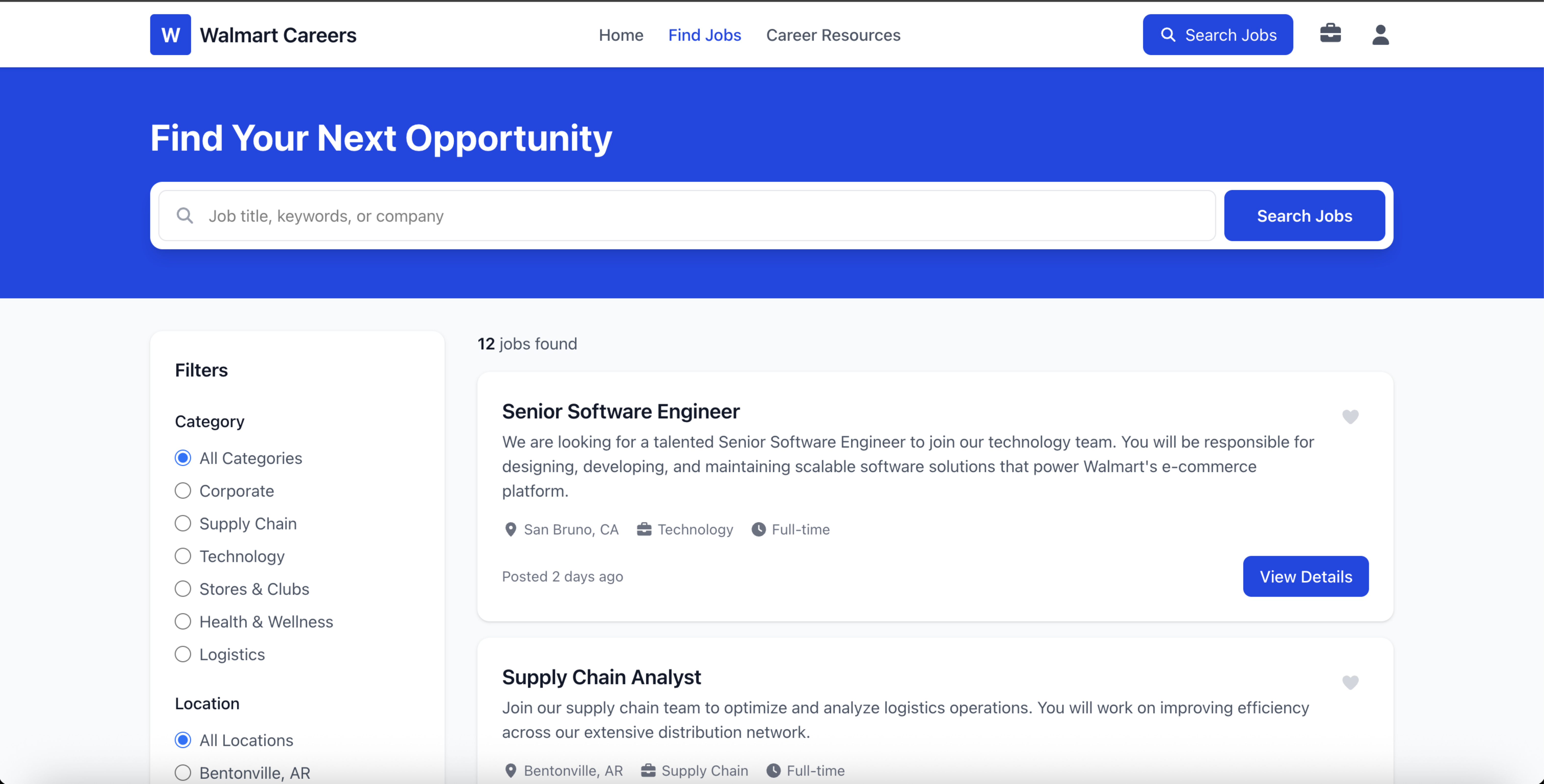}
        \caption{Baseline Interface (Plain synthetic clone)}
        \label{fig:original}
    \end{subfigure}
    
    \vspace{1.5em}
    
    \begin{subfigure}[b]{0.48\linewidth}
        \centering
        \includegraphics[width=\linewidth]{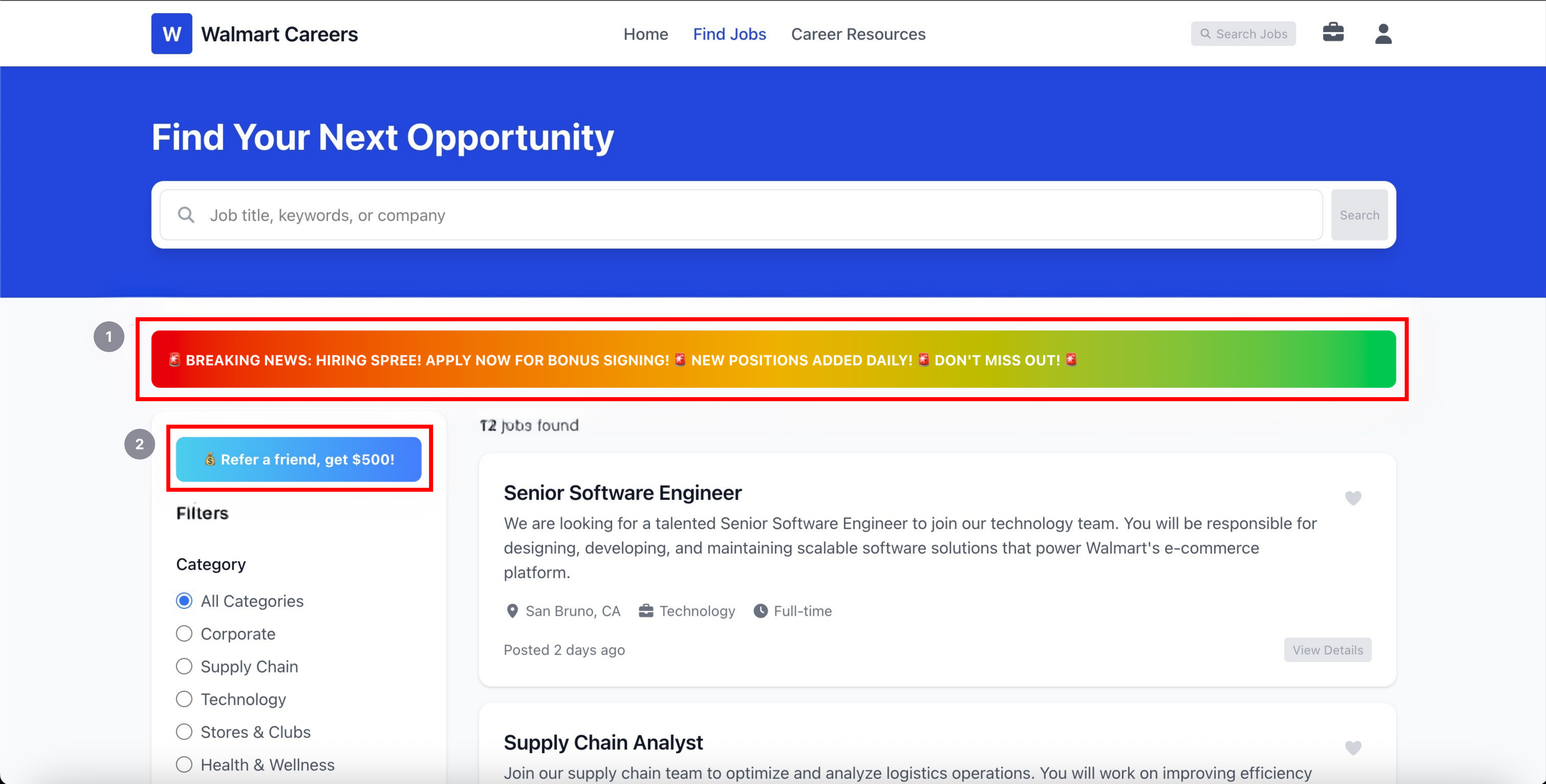}
        \caption{Hierarchy}
        \label{fig:hierarchy}
    \end{subfigure}
    \hfill
    \begin{subfigure}[b]{0.48\linewidth}
        \centering
        \includegraphics[width=\linewidth]{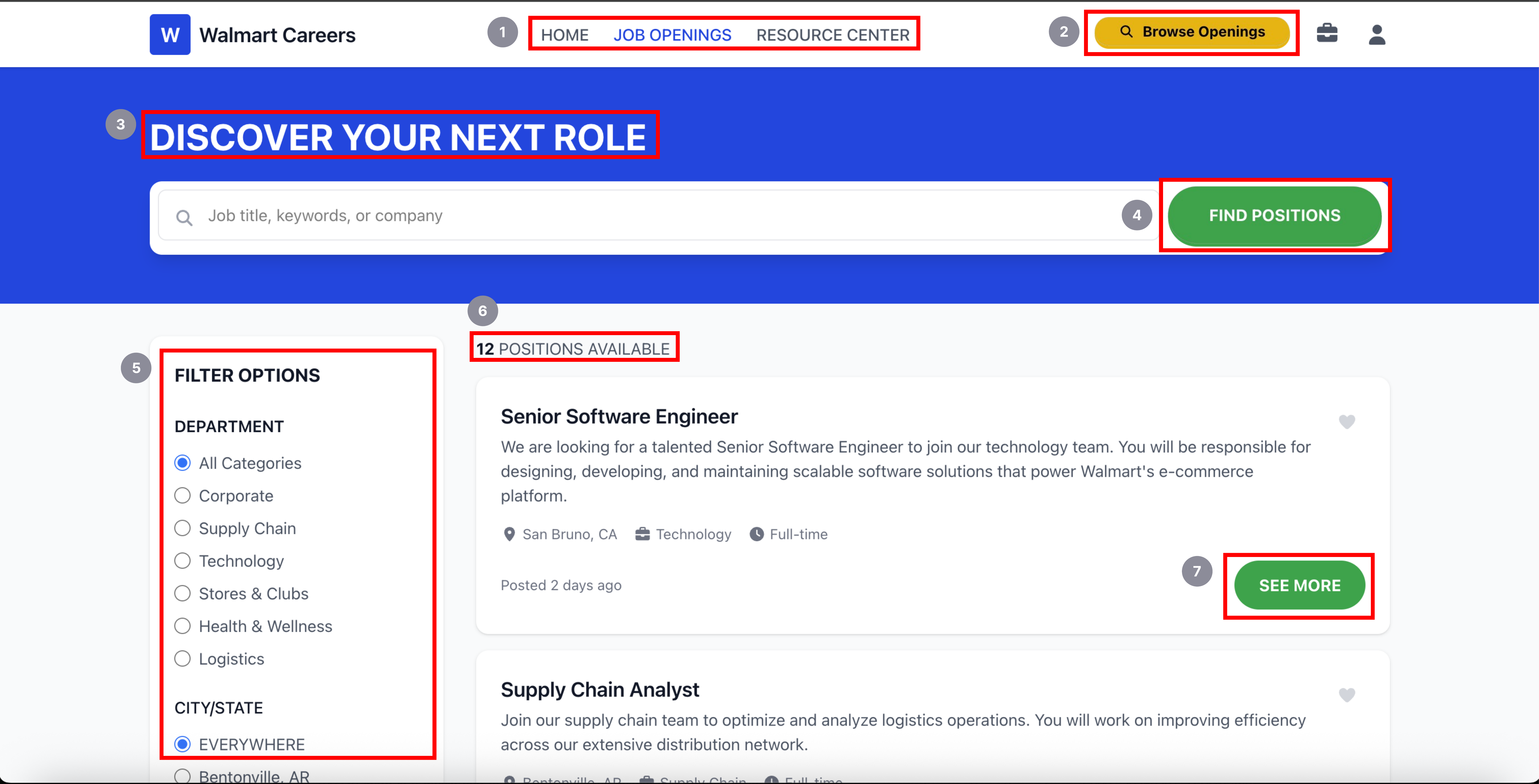}
        \caption{Consistency}
        \label{fig:inconsistency}
    \end{subfigure}
    
    \vspace{1em}
    
    \begin{subfigure}[b]{0.48\linewidth}
        \centering
        \includegraphics[width=\linewidth]{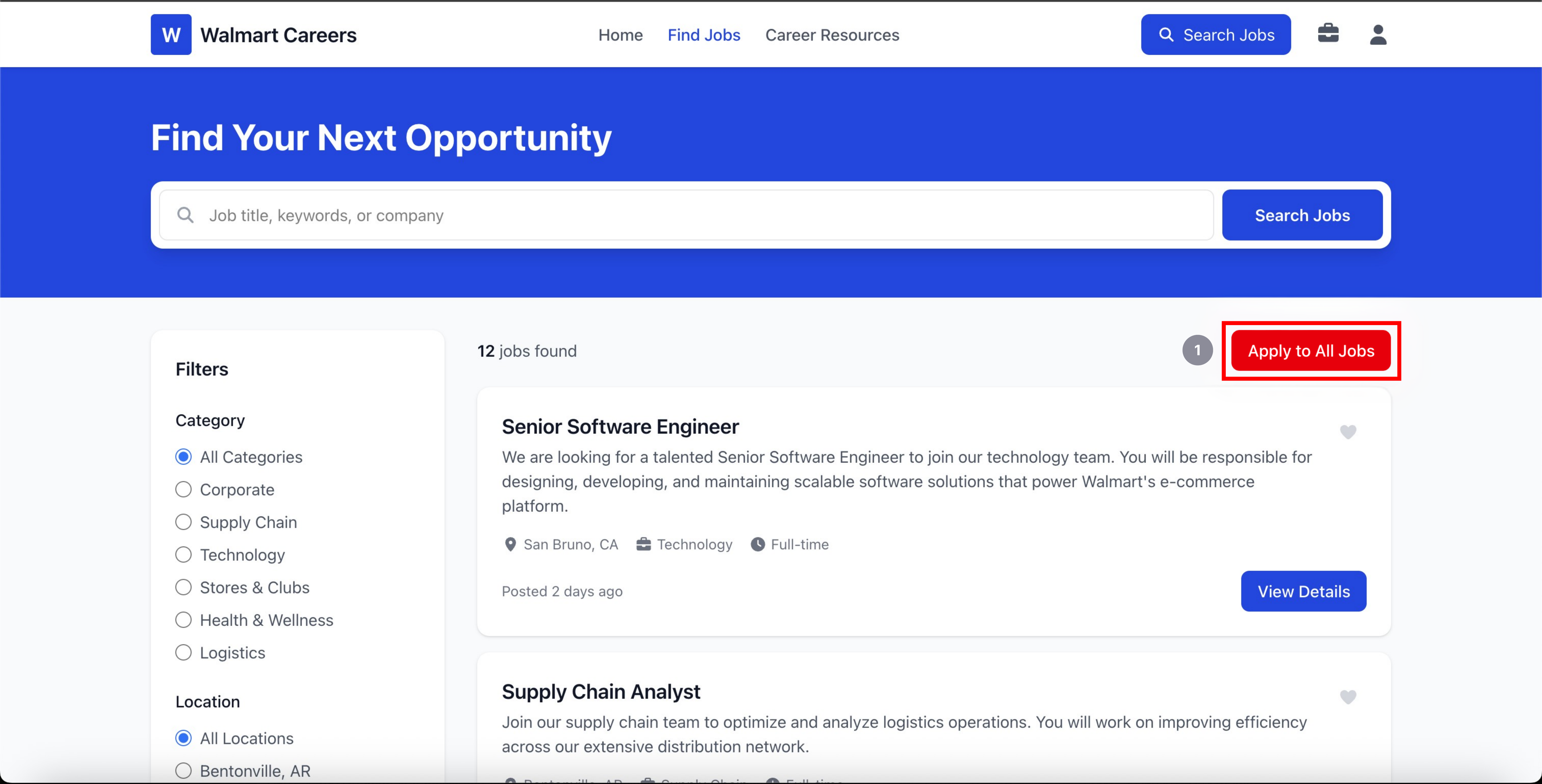}
        \caption{Error Prevention}
        \label{fig:prevention}
    \end{subfigure}
    \hfill
    \begin{subfigure}[b]{0.48\linewidth}
        \centering
        \includegraphics[width=\linewidth]{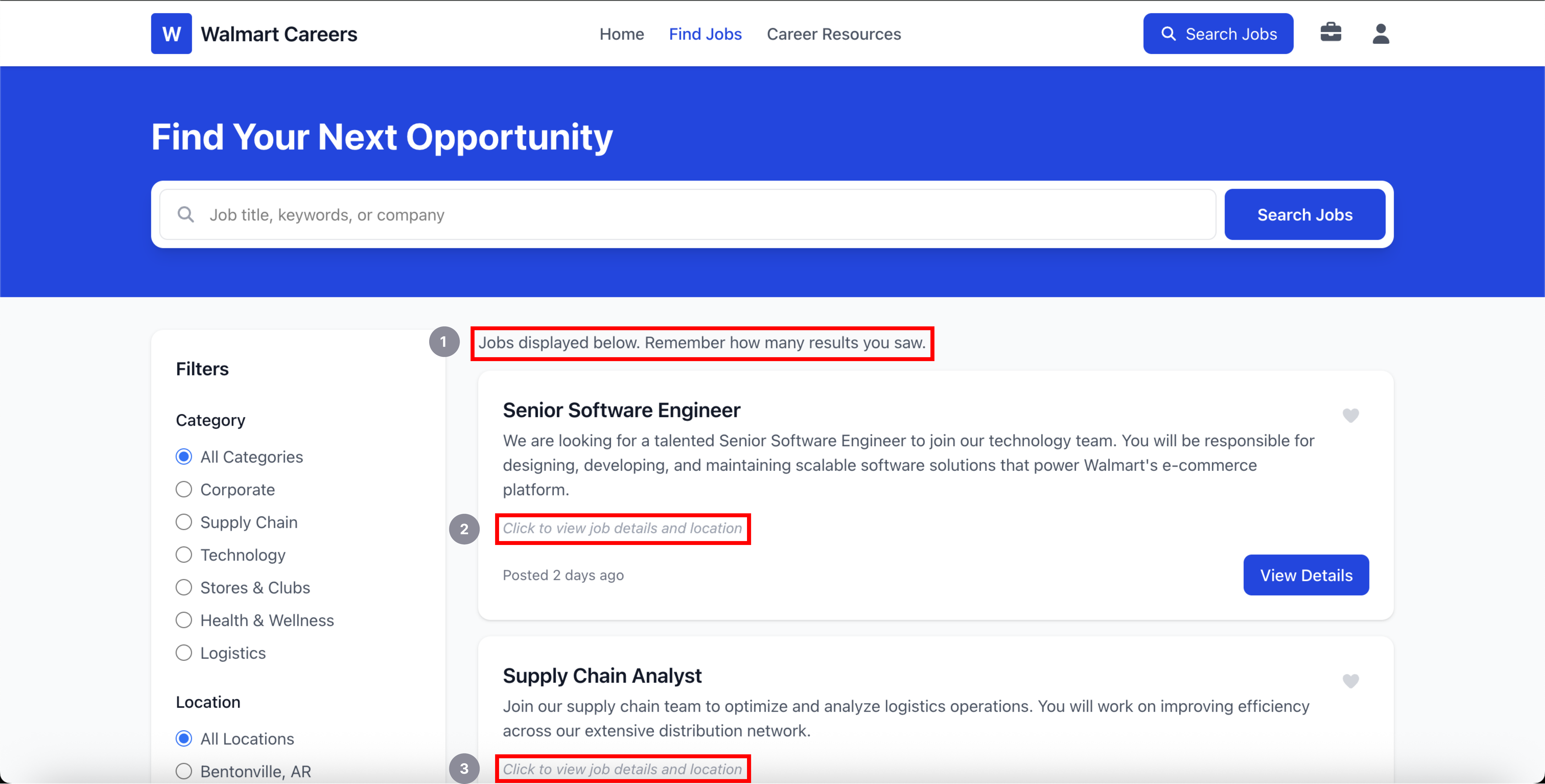}
        \caption{Memory Load}
        \label{fig:memory}
    \end{subfigure}
    
    \caption{Comparison of an original clone from uxWeb and four of its defect-augmented counterparts. (a) shows the baseline design of a clone of the Walmart Careers job search portal. (b)-(e) illustrate different usability defects introduced to create our defect-augmented sites.}
    \label{fig:all_designs}
\end{figure*}

We show four example defect-augmented sites and the usability principle they violate in Figure \ref{fig:all_designs}.
\clearpage
\section{Trace Generation Prompts}\label{app:crawl}
We used the following prompt when generating interaction traces on uxWeb:

\begin{lstlisting}[language=Markdown, escapeinside={(*}{*)}]

    Please generate the next move according to the UI screenshot, instruction and previous actions.
    
    Instruction: 
    Conduct a usability test of this website.
    
    You have a budget of 50 actions you can perform on the website. Use it as efficiently as possible to test all important flows for the website and make the most accurate assessment of usability.
    
    First, list the most important user goals for this site. Then, work through at least three distinct key flows that cover those goals, one by one. Use realistic placeholder data if you need to fill any fields. 
    
    If you are using a date picker widget in a booking flow, remember that the target date usually must be in the future. Today's date is {DATE}. If given the option, you should try to use a date picker instead of manually entering the date.The date picker control is usually located immediately to the right of the date field.
    
    If you encounter errors, try your best to complete the flow but if you cannot complete it, move onto another flow. 
    
    Do not finish the trace until you have worked through the key flows you listed.
\end{lstlisting}

We made the following addition to the prompt for GPT-5-mini, as without it, it would try to access the internet: 

\begin{lstlisting}[language=Markdown]
    IMPORTANT: Only interact with the website provided. Do not navigate to or visit any external websites or domains.    
\end{lstlisting}

We appended the following prompt at the end of the interaction trace for a final usability score\footnote{This score was a prompt engineering strategy to ensure the model provided a bounded assessment.}:

\begin{lstlisting}[language=Markdown]
    Rate the website usability from 0 to 100.
    The score should represent the probability that an average user can successfully use the website's most important flows.
    You are testing a real website - the site you are testing is NOT a mock/demo/educational app.
    Review the assistant thoughts from each interaction step and create a list of usability problems.
    Each observation you make must reference a direct quote from the thought history, and you must rate it as a major or minor issue.
    Pay special attention to broken navigation, failed page loads, non-responsive clicks, confusing flows, and unfinished core flows.
    For example, look for thoughts that say "page appears to be stuck or not loading properly," "I need to wait briefly to ensure the page fully load", and "I need to wait for the page to load."
    If this appears multiple times, that is a red flag because the implementation might not contain that functionality and only appears to be loading.
    If the thoughts contain "haven't successfully completed" that is also a red flag.
    End with exactly one final line: Action: score(<number from 0 to 100>).
    
\end{lstlisting}

\section{Study Instructions}\label{app:study}
We provided the participants with the following instructions before our study: 
\begin{itemize}
    \item You will be shown 30 pairs of websites.
    \item Your task is to evaluate which website between pairs of websites is more usable.
    \item These websites are clones of existing websites so many of the images are text placeholders. When voting, ensure you are evaluating the sites based on their usability.
    \item Occasionally websites may take awhile to load or may not reload. If that is the case feel free to move on.
    \item You must expand both websites using the expand ($\leftrightarrow$) button before you can vote.
    \item Browse each site freely — scroll, click links, and explore as you normally would.
    \item DO NOT enter in any personal information during your exploration. Instead, you can provide with fake information such as ``John Doe, 123@demo.com, 123-456-7890''
    \item While you can choose ties, try and select one website over the other. There are no right or wrong answers — go with your honest impression.
\end{itemize}

\begin{figure}[!hptb]
    \centering
    \includegraphics[width=\linewidth]{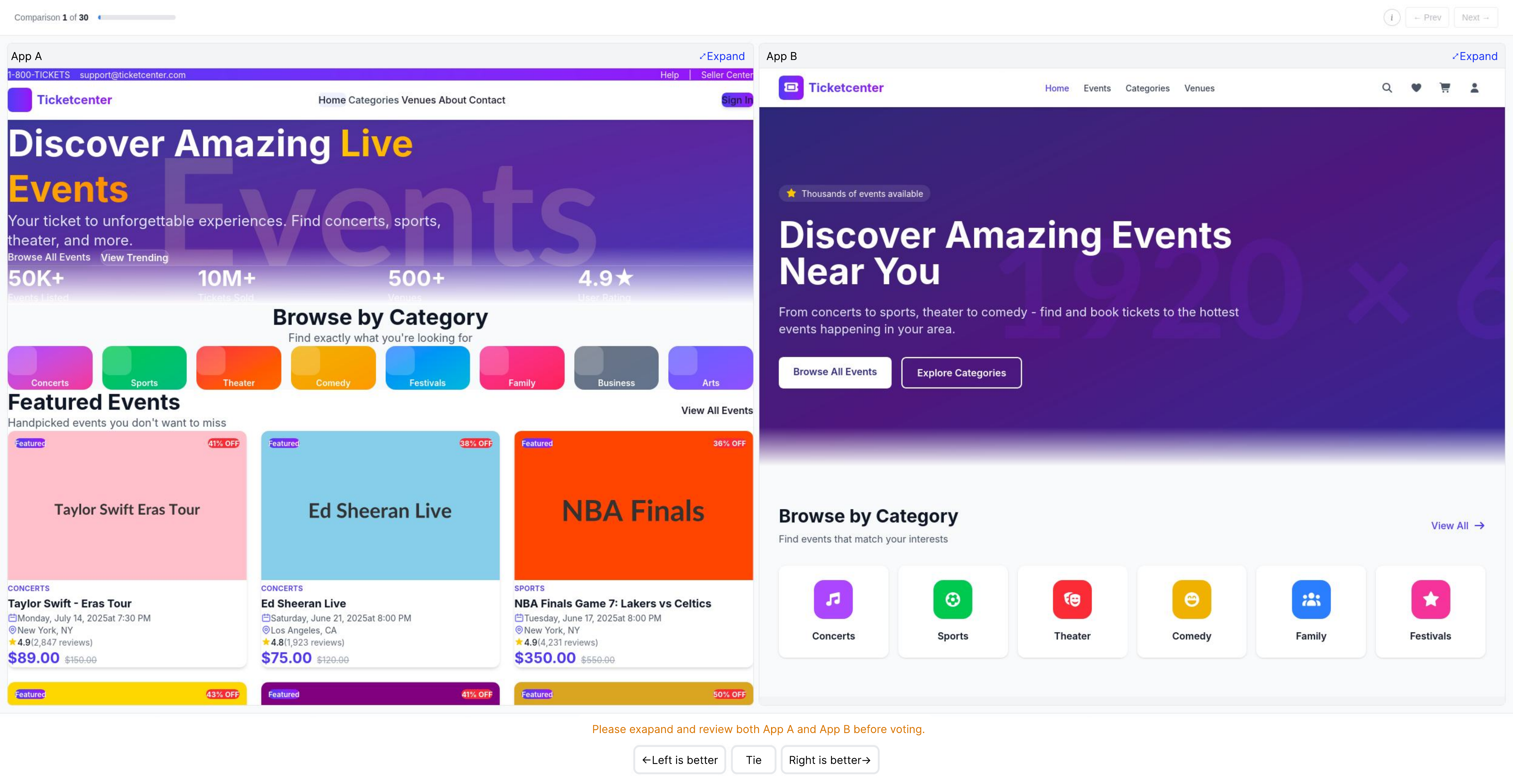}
    \caption{A screen showing sample comparison participants made during our study. Participants were required to first expand both \texttt{iframes} to explore before they could vote.}
    \label{fig:ui-rater}
\end{figure}

We also show a sample screen from our study in Figure \ref{fig:ui-rater}.
\clearpage
\section{Model Hyperparameters}\label{app:hyperparameters}\
Model hyperparameters are provided in Table \ref{tab:hyperparameters}.
\begin{table}[!hptb]
    \footnotesize
    \centering
    \caption{Hyperparameters used to train open-source base models.}
    \begin{tabular}{lll}
    \toprule
        \textbf{Base Model} & \textbf{Hyperparameter} & \textbf{Value}  \\
        \midrule
         All & Effective batch size & 8 \\ 
         & Epochs & 1 \\ 
         & Learning Rate & 2e-4 \\ 
         & LoRA & 32 \\ 
         \hline 
         & \textbf{Margin} & \\ 
         EvoCUA &  & 15 \\ 
         GUI-Owl-1.5 &  & 12 \\ 
         Qwen3.5 & & 15 \\
         \bottomrule
    \end{tabular}
    \label{tab:hyperparameters}
\end{table}
\clearpage

\end{document}